\documentclass[letterpaper, 10 pt, conference]{ieeeconf}  

\IEEEoverridecommandlockouts                              

\usepackage{graphics} 
\usepackage{epsfig} 
\usepackage{times} 
\usepackage{amsmath} 
\usepackage{amssymb}  
\usepackage{microtype}
\usepackage{bm}%
\usepackage{cite}

\usepackage{epsfig}
\usepackage{graphicx}
\usepackage{float}
\usepackage{siunitx}
\usepackage{verbatim}
\usepackage{subcaption} 
\usepackage{booktabs}
\usepackage{cuted}
\usepackage{capt-of}
\usepackage{url}
\usepackage{arydshln}
\usepackage{bm}
\usepackage[noend]{algpseudocode}
\usepackage{algorithmicx,algorithm}
\usepackage{multirow}
\usepackage{lipsum}

\usepackage{booktabs,arydshln}

\makeatletter
\def\adl@drawiv#1#2#3{%
        \hskip.5\tabcolsep
        \xleaders#3{#2.5\@tempdimb #1{1}#2.5\@tempdimb}%
                #2\z@ plus1fil minus1fil\relax
        \hskip.5\tabcolsep}
\newcommand{\cdashlinelr}[1]{%
  \noalign{\vskip\aboverulesep
           \global\let\@dashdrawstore\adl@draw
           \global\let\adl@draw\adl@drawiv}
  \cdashline{#1}
  \noalign{\global\let\adl@draw\@dashdrawstore
           \vskip\belowrulesep}}
\makeatother

\title{\LARGE \bf
Video Deblurring by Fitting to Test Data
}

\author{Xuanchi Ren*, Zian Qian*, and Qifeng Chen
}

\begin{document}


\twocolumn[{%
\renewcommand\twocolumn[1][]{#1}%
\maketitle
\begin{center}
\vspace{-1.5 em}
\renewcommand\arraystretch{0.5} 
\centering
\begin{tabular}{@{}c@{\hspace{0.5mm}}c@{\hspace{0.5mm}}c@{\hspace{0.5mm}}c@{\hspace{0.5mm}}c@{}}
&\includegraphics[width=0.24\linewidth]{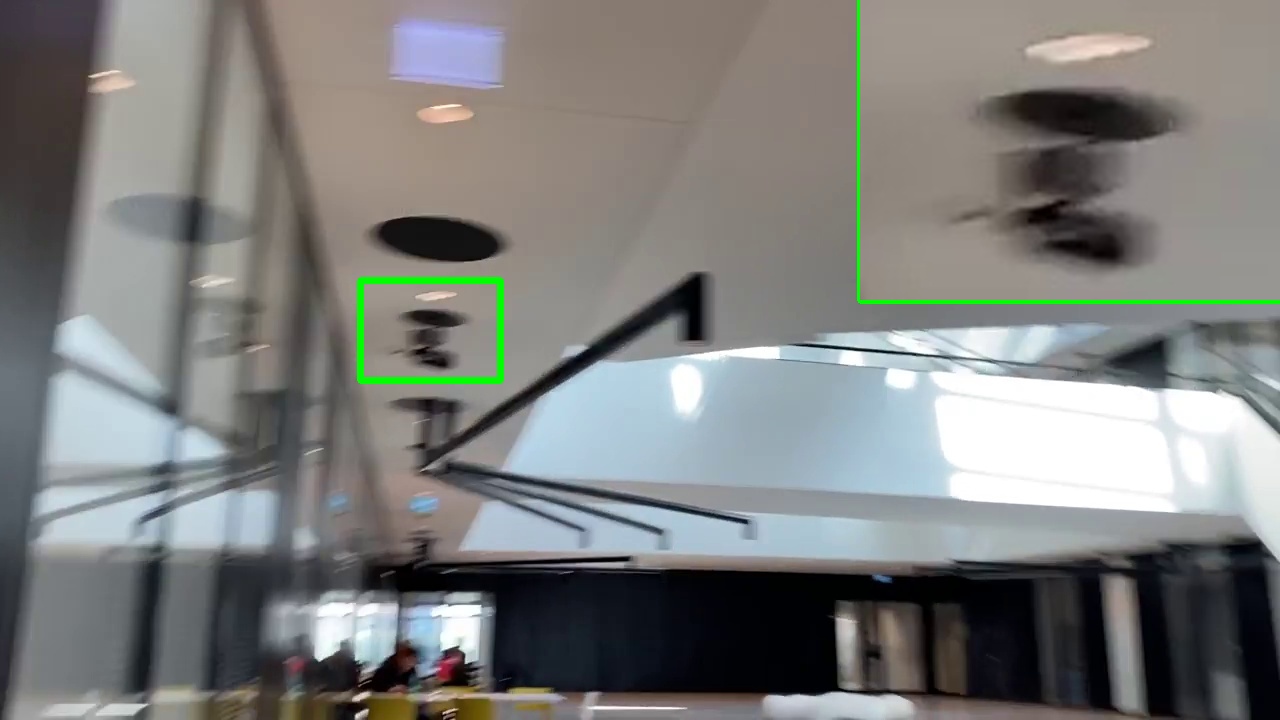}&
\includegraphics[width=0.24\linewidth]{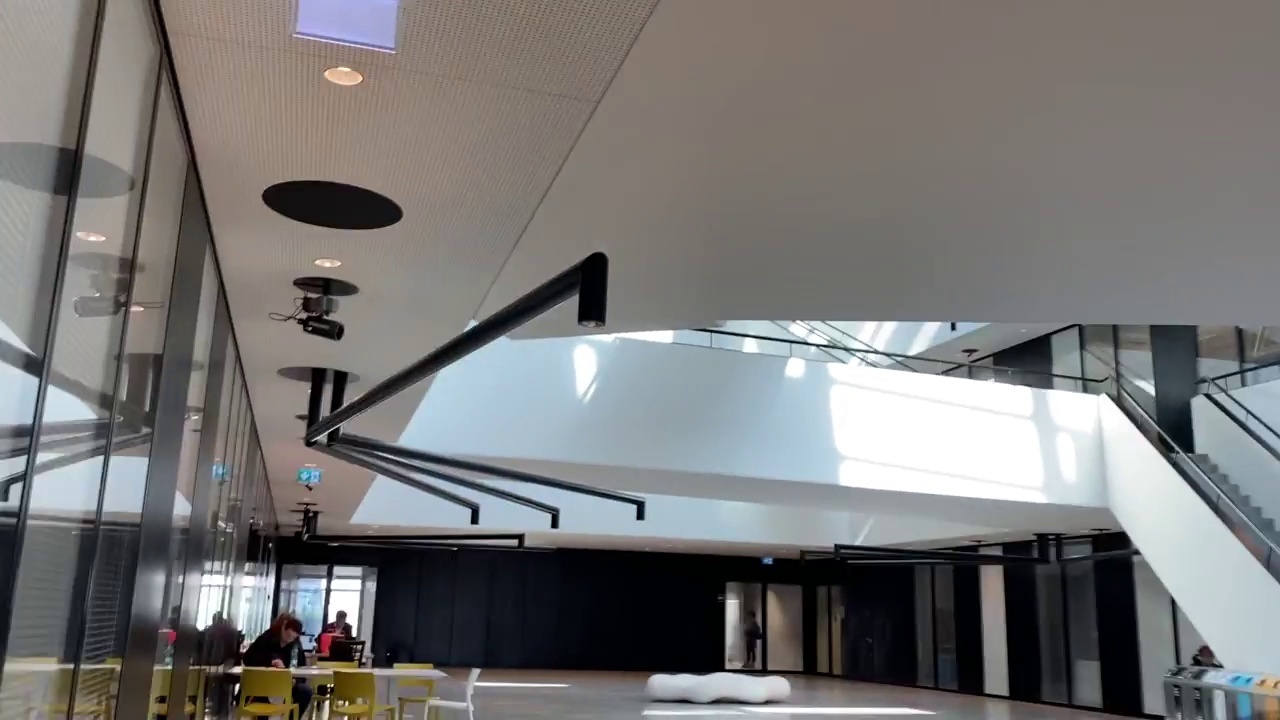}&
\includegraphics[width=0.24\linewidth]{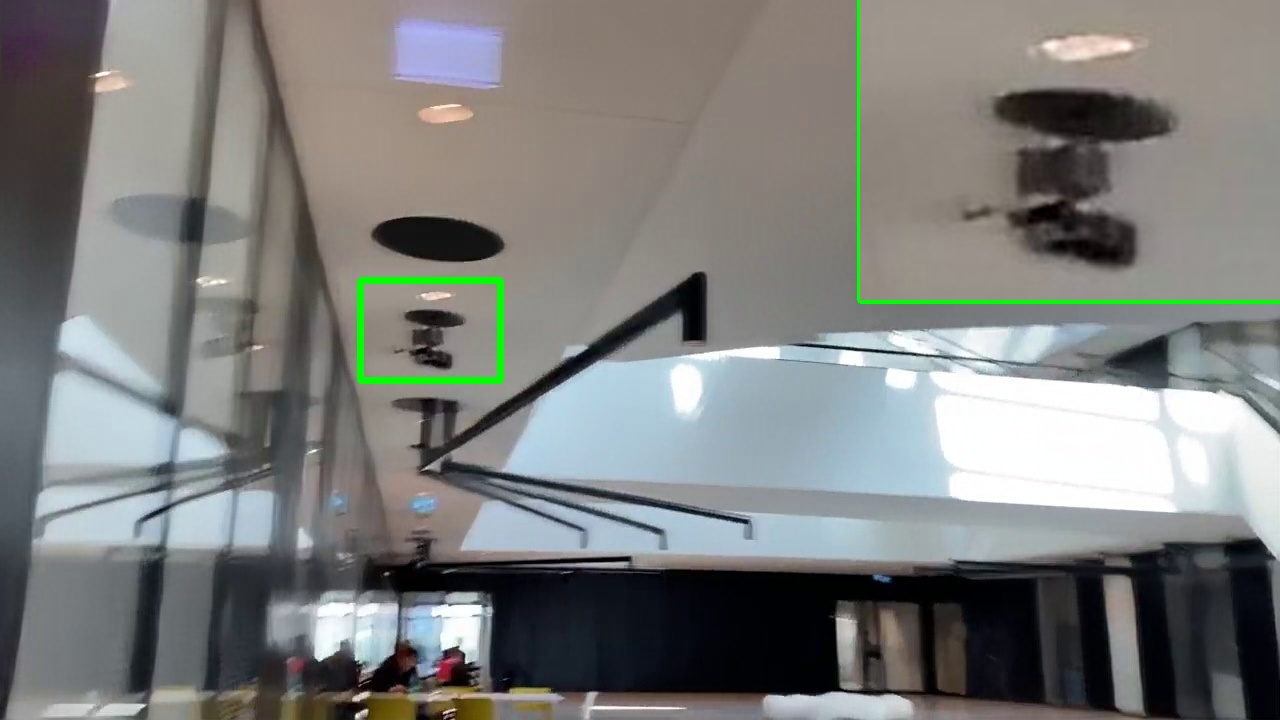}&
\includegraphics[width=0.24\linewidth]{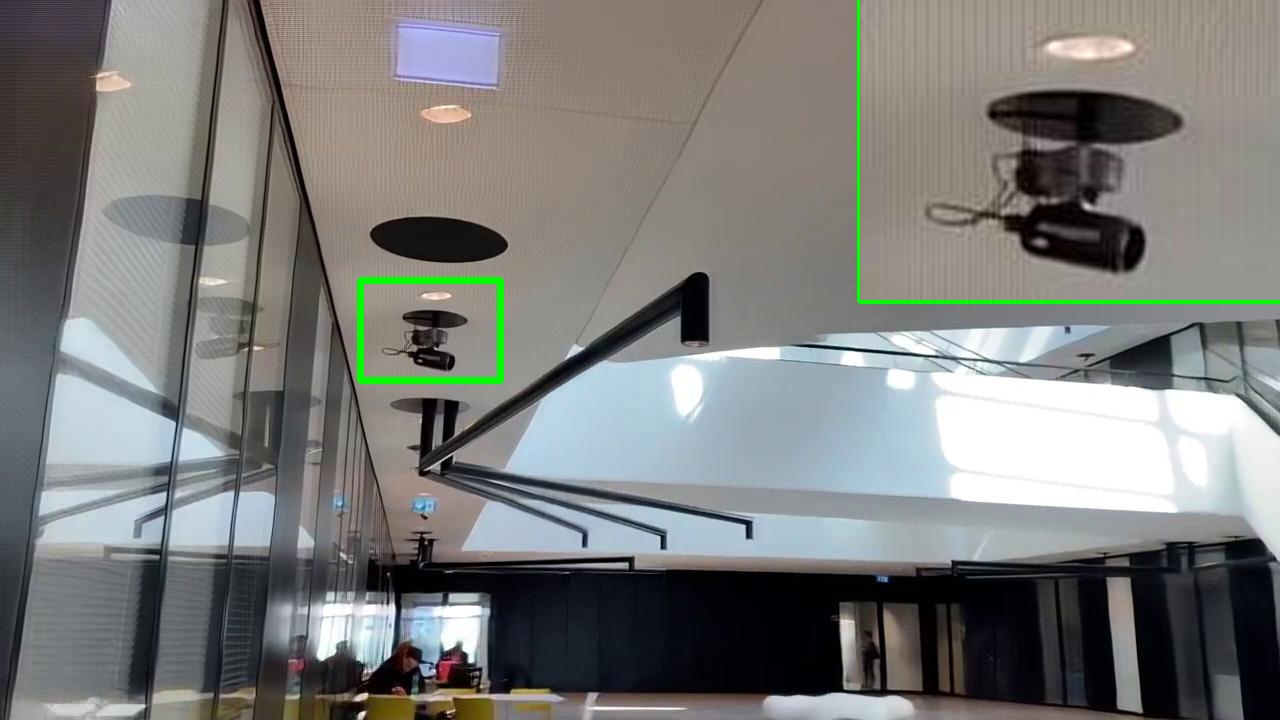}\\

&\includegraphics[width=0.24\linewidth]{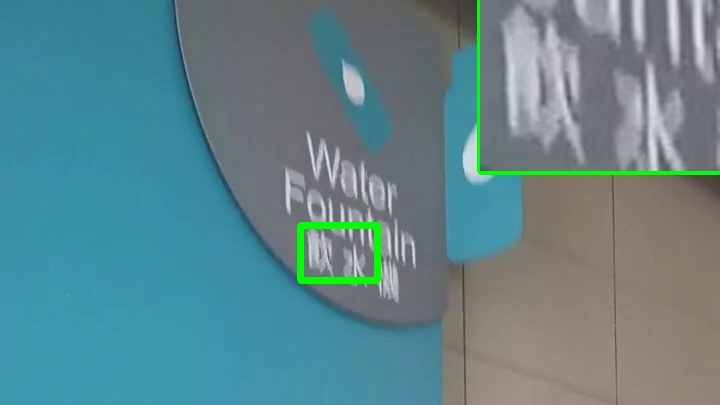}&
\includegraphics[width=0.24\linewidth]{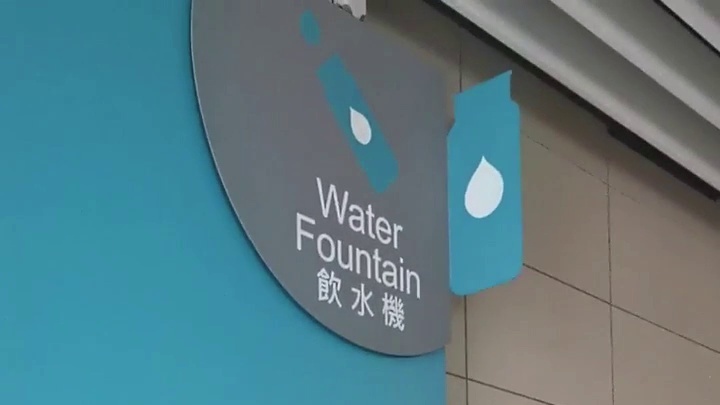}&
\includegraphics[width=0.24\linewidth]{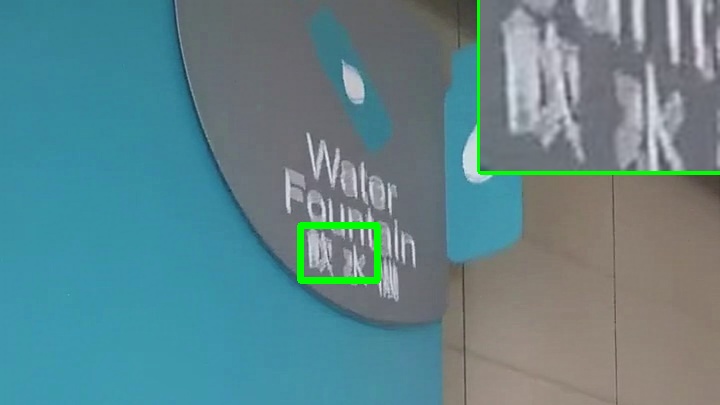}&
\includegraphics[width=0.24\linewidth]{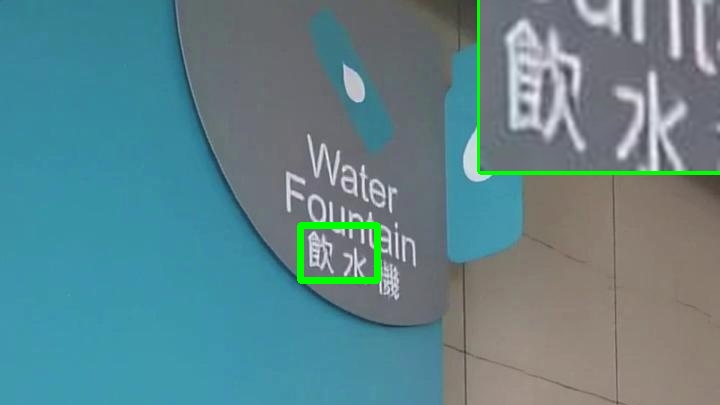}\\

&\includegraphics[width=0.24\linewidth]{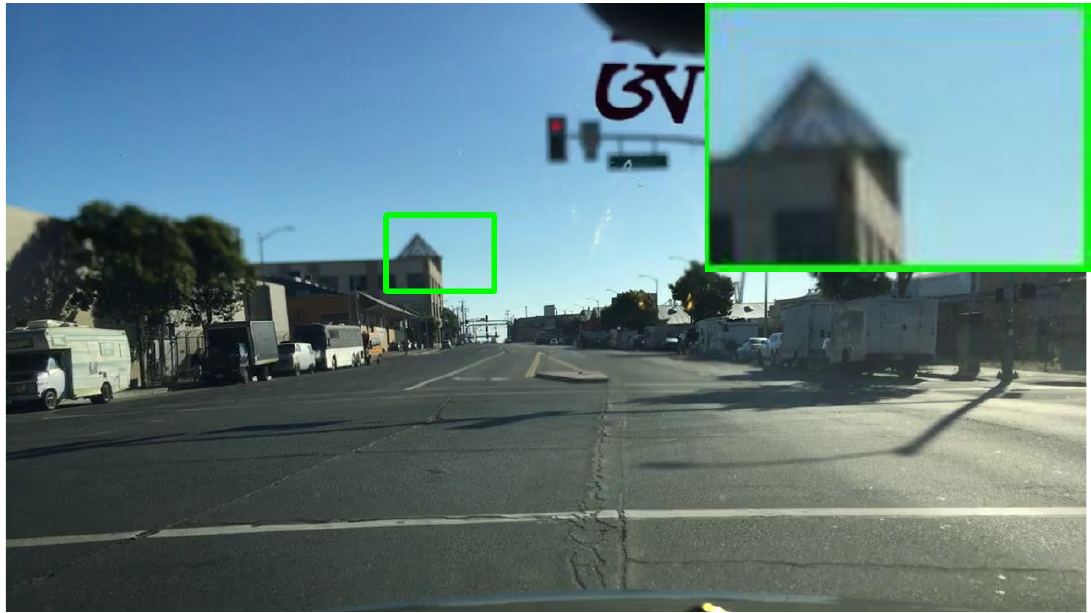}&
\includegraphics[width=0.24\linewidth]{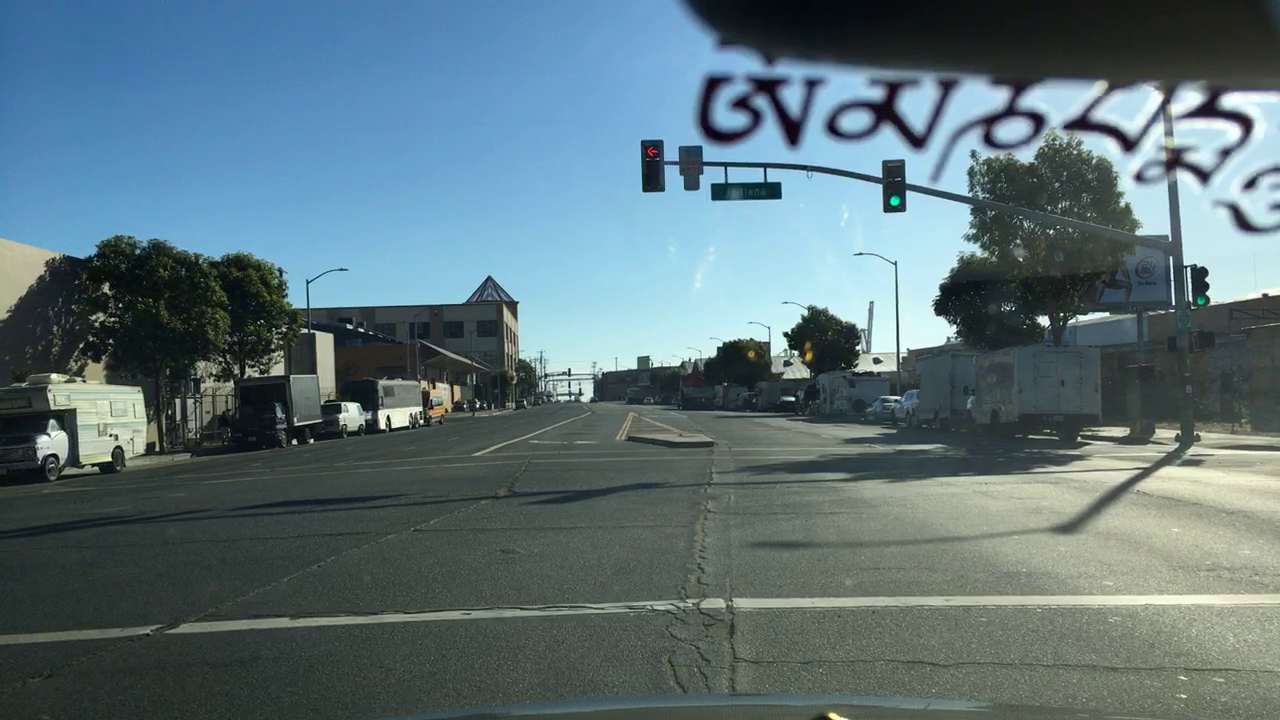}&
\includegraphics[width=0.24\linewidth]{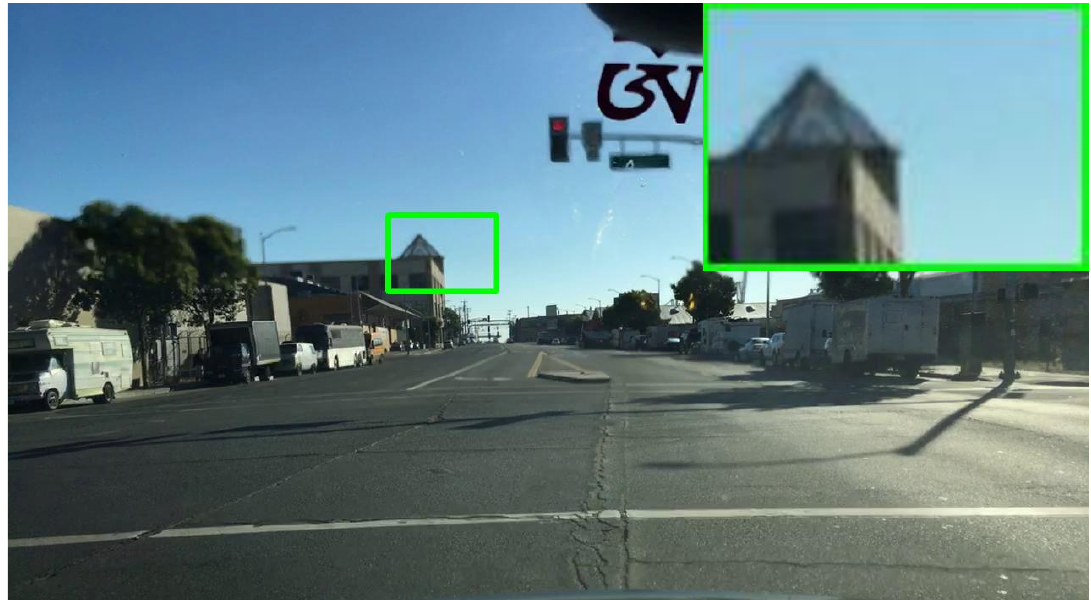}&
\includegraphics[width=0.24\linewidth]{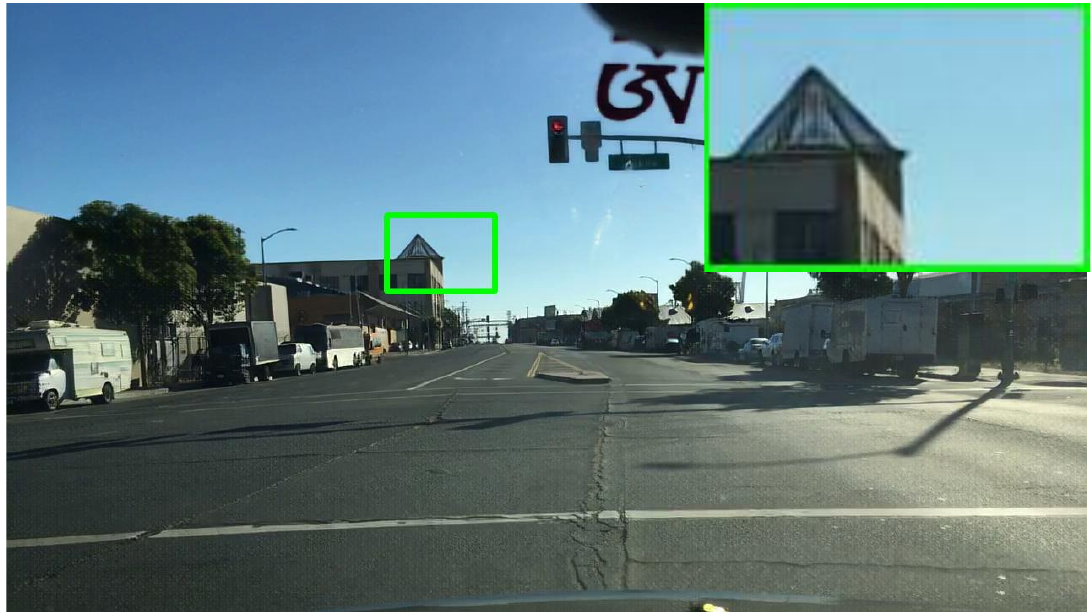}\\
&(a) Blurry frame &(b) Sharp frame &(c) DeblurGAN-v2 &(d) Ours \\
\end{tabular}
\captionof{figure}{(a) and (b) are two frames from the same video. Note that the sharp textures in (b) can be used to refine (a). (c) is the video deblurring result by a state-of-the-art deblurring method proposed by Kupyn et al.~\cite{kupyn2019deblurgan} and (d) is our result.}
\label{fig:observation}
\end{center}
}]

{
  \renewcommand{\thefootnote}%
    {\fnsymbol{footnote}}
  \footnotetext[1]{Co-first authors. Xuanchi Ren, Zian Qian, and Qifeng Chen (cqf@ust.hk) are with the Department of Computer Science and Engineering, HKUST.}
}

\begin{abstract}

Motion blur in videos captured by autonomous vehicles and robots can degrade their perception capability. In this work, we present a novel approach to video deblurring by fitting a deep network to the test video. Our key observation is that some frames in a video with motion blur are much sharper than others, and thus we can transfer the texture information in those sharp frames to blurry frames. Our approach heuristically selects sharp frames from a video and then trains a convolutional neural network on these sharp frames. The trained network often absorbs enough details in the scene to perform deblurring on all the video frames. As an internal learning method, our approach has no domain gap between training and test data, which is a problematic issue for existing video deblurring approaches. The conducted experiments on real-world video data show that our model can reconstruct clearer and sharper videos than state-of-the-art video deblurring approaches.
\end{abstract}


\begin{figure*}[t]
\centering
\includegraphics[width=\linewidth]{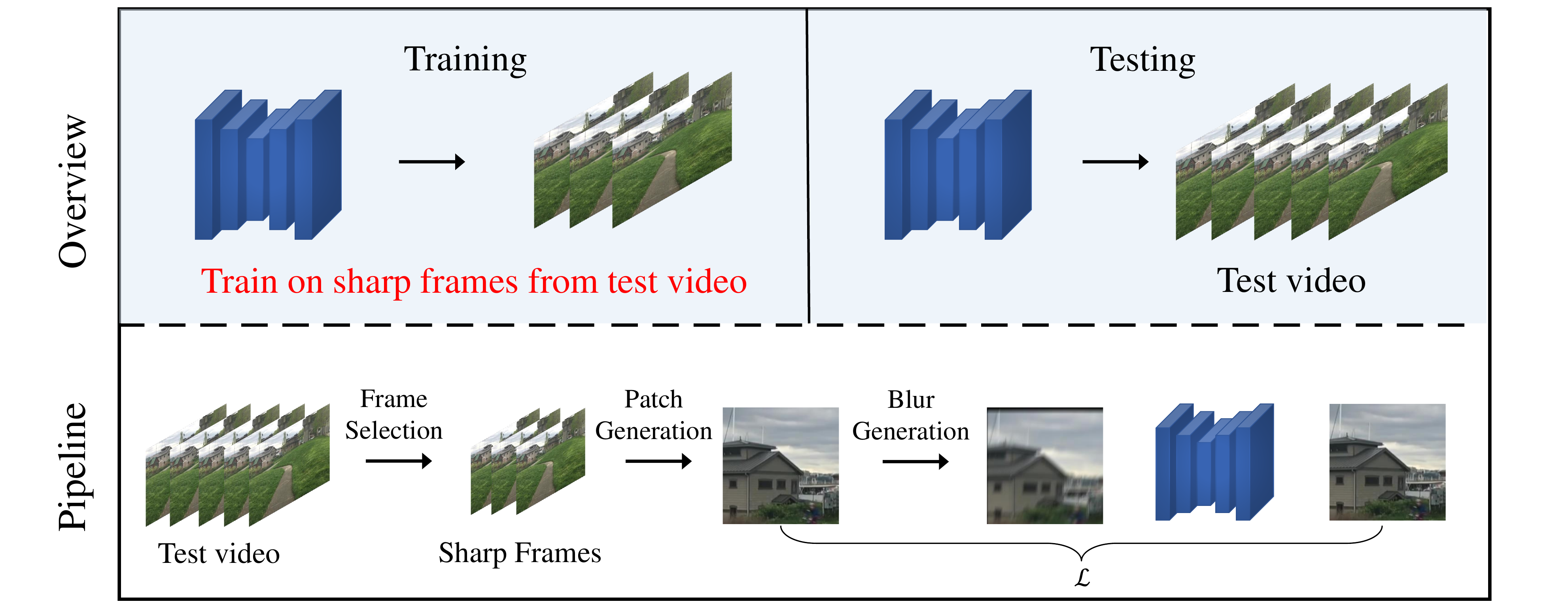}
\caption{Overview of our fitting-to-test-data pipeline. In our pipeline, for each test video, we fit a deep learning model to it and obtain a test-data-aware model.}
\label{fig:pipeline}
\vspace{0mm}
\end{figure*}

\section{INTRODUCTION}

For autonomous robots, videos are often captured while the cameras are moving and thus motion blur is an inevitable visual artifact in the captured video. Motion blur is mainly caused by the fast movement of the camera or moving objects in the environment. The motion blur in videos captured by self-driving cars is often severe, which may hinder its perception performance. For instance, a dashcam video may contain critical information about a car plate in some frames that appear to be blurry, and a user would want to recover the underlying details. 

Recovering sharp frames from a blurry video is still challenging for learning-based methods due to the domain gap between training and testing data. For example, a model trained on a dataset captured by a dashcam in the streets may have bad performance when tested on videos captured by a drone.
Moreover, since it is hard to capture the corresponding sharp and blurry real-world videos, researchers synthesize blurry videos by averaging sharp videos at a high frame rate. 
As illustrated in Figure \ref{fig:observation} (c), the model trained on synthetic data in a different scenario fails to handle the motion blur in a real-world video in daily life. Therefore, the state-of-the-art video deblurring methods often produce poor results on real-scene videos.

\begin{figure*}[t]
\centering
\begin{tabular}{@{}c@{\hspace{1mm}}c@{\hspace{1mm}}c@{\hspace{1mm}}c}
\includegraphics[width=0.24\linewidth]{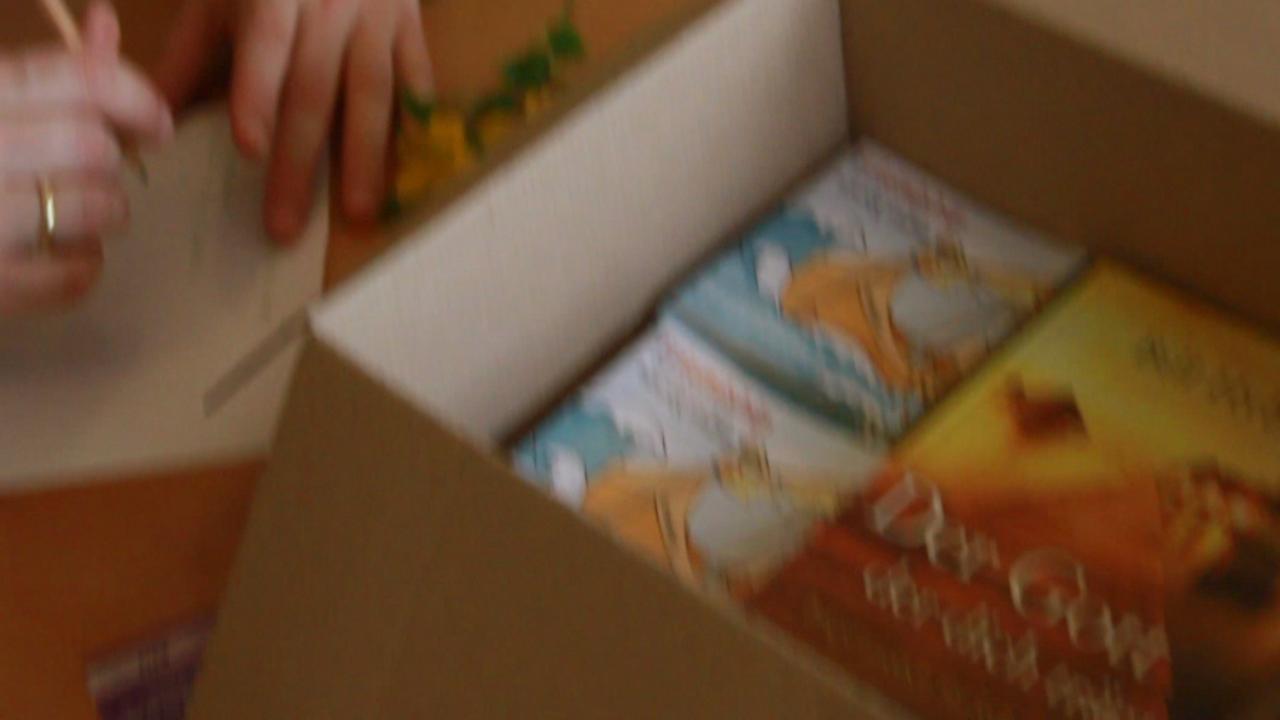} &
\includegraphics[width=0.24\linewidth]{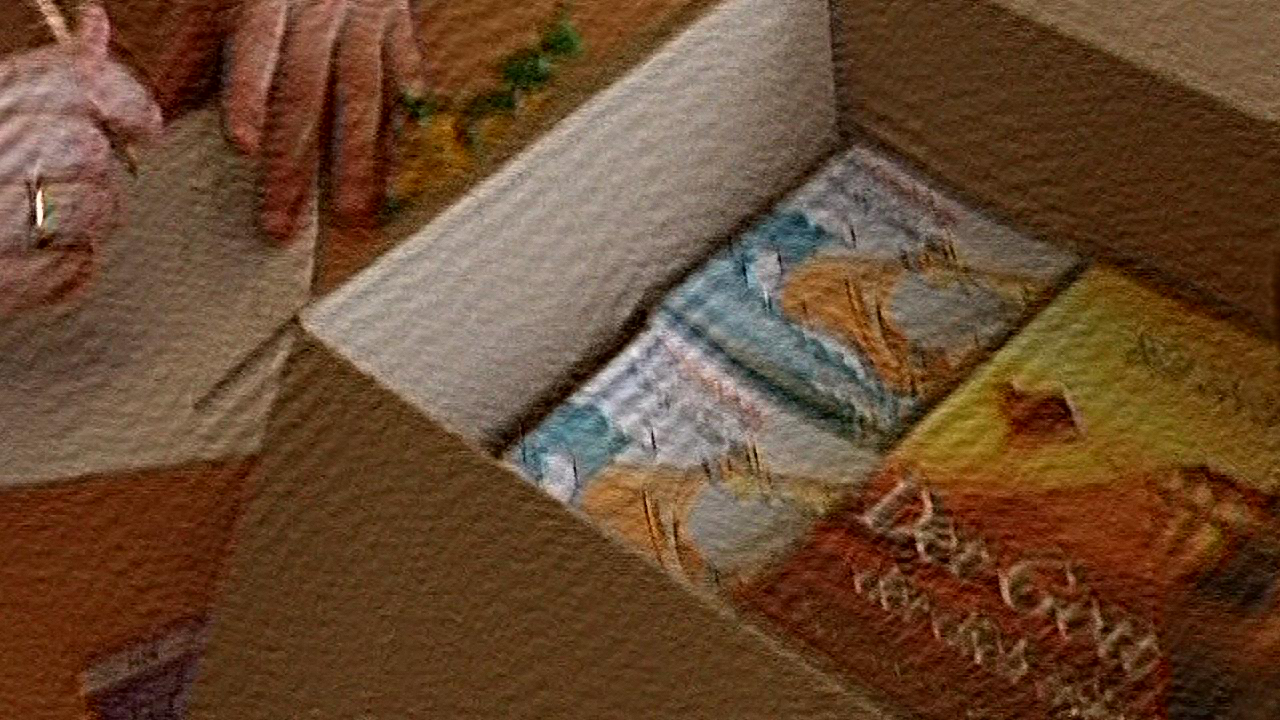} &
\includegraphics[width=0.24\linewidth]{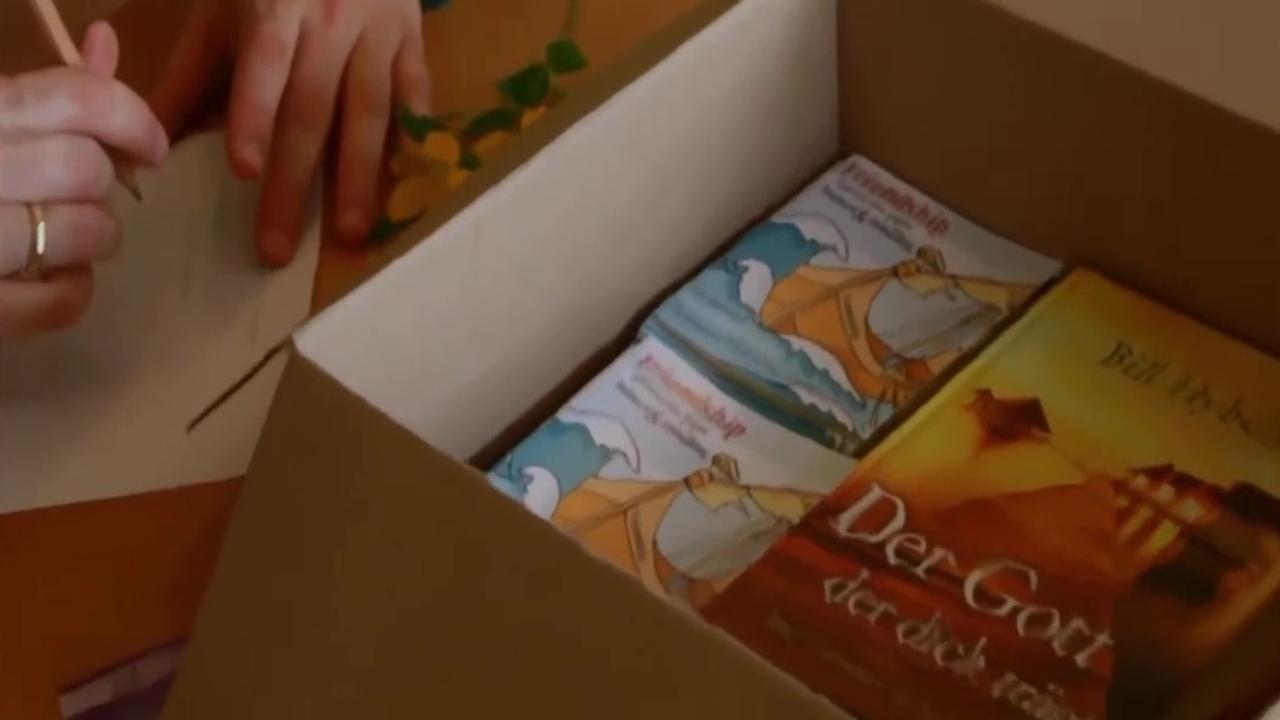} &
\includegraphics[width=0.24\linewidth]{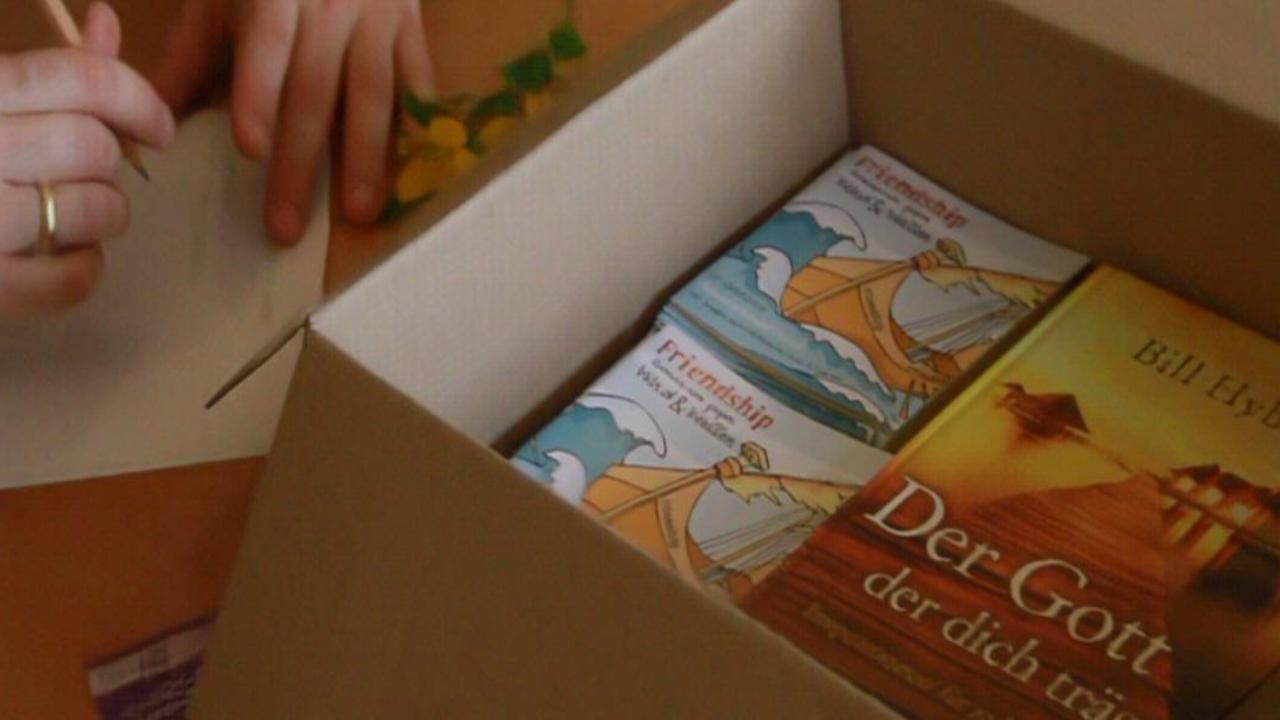}\\
Input& SelfDeblur~\cite{ren2019neural}& Cho et al.~\cite{cho2012video} & Ours \\
\end{tabular}
\caption{The visual comparison between ours and two internal deblurring methods. The result of Cho et al.~\cite{cho2012video} is extracted from their demo video. A slight shift is caused by their stabilization. Zoom in for details.}
\label{fig:TOG}
\end{figure*}

Researchers have proposed various ideas to address the domain gap issue for video deblurring, but their actual performance is still not satisfactory yet. Kupyn et al.~\cite{KupynBMMM18,kupyn2019deblurgan} improved the generalization ability of the pre-trained model by estimating the residual between blurry and clear frames. As illustrated in Figure~\ref{fig:observation}, the results are still far from perfect in general scenes. Ulyanov et al.~\cite{DIP} demonstrated that high-frequency information was generated in the posterior phase of the training process. This property forms the basis of internal learning for video deblurring. Ren et al.~\cite{ren2019neural} and Asim et al.~\cite{asim2019blind} estimated blur kernels frame by frame with deep image prior~\cite{DIP}. While these approaches precluded the problem of the domain gap, they still relied on handcrafted prior and had poor performance on real-world blur, as shown in Figure~\ref{fig:TOG}.

To further address the domain gap problem, we introduce a self-supervised pipeline by fitting a deep learning model only on the test video. As shown in Figure~\ref{fig:pipeline}, unlike traditional supervised pipeline, our method does not rely on training on a large (synthetic) dataset. Our approach is built upon an observation that some frames in a video with motion blur are extremely sharp and clear, as illustrated in Figure~\ref{fig:observation}. Based on this crucial observation, we exploit the internal information of a video by training a simple but efficient convolutional network (CNN) on the sharp-blurry patch pairs generated by our blur generation strategy. As such, our test-data-aware CNN model can adapt to any scenarios and settings of daily videos without the domain gap between training and test data. We also improve our training process with better initialization using MAML~\cite{finn2017model}, which successfully reduces the running time for a video from hours to 5 minutes.

To analyze the performance of our method, we conduct an extensive user study on Amazon Mechanical Turk. We compare our method with several state-of-the-art image and video deblurring approaches on a real-world dataset, and the result shows our method significantly outperforms the three approaches. Moreover, our accelerated version also achieves state-of-the-art performance in video deblurring while its running time is substantially reduced by two orders of magnitude.

In summary, the contributions of our work are:
\begin{itemize}
\renewcommand{\labelitemi}{\textbullet}
\setlength{\itemsep}{0pt}
\item We propose a video deblurring pipeline without the need for a large training dataset by fitting a deep network to the sharp frames in a test video.
\item With our blur generation strategy and loss reweighting trick, our pipeline significantly outperforms existing state-of-the-art video deblurring methods, as demonstrated in the user study. 
\item Combined with meta-learning, our pipeline can be accelerated by about two orders of magnitudes, with little sacrifice in video deblurring quality, but also achieves state-of-the-art performance.
\item We collected a dataset containing 70 real-world videos with motion blur that can be used for evaluation on the deblurring task and will be made available to the public. 
\end{itemize}

\section{Related Work}

\textbf{Blind deblurring.} 
Early work on image deblurring is often based on manually designed prior by jointly estimated a blur kernel and the underlying sharp image via deconvolution~\cite{kundur1996blind,KrishnanTF11,chang2014new}. 
Meanwhile, classic work on video blurring proposed to aggregate neighboring frames and transferred the sharp pixels to the central frames~\cite{cho2012video}. However, such a traditional method can not handle large depth variations.

Recently, the development of deep learning brings breakthrough to image deblurring task~\cite{KupynBMMM18, nah2017deep, purohit2019region, tao2018srndeblur, ZhangPRSBL018}. Nah et al.~\cite{nah2017deep} used an end-to-end multi-scale CNN to restore the images. Kupyn et al.~\cite{kupyn2019deblurgan} extended the GAN-based method from ~\cite{KupynBMMM18} to Feature Pyramid Network with double-scale discriminator. Lately, Purohit et al.~\cite{purohit2019region} combined attention modules to discover the non-local spatial relationships inside an image for deblurring.

Based on these image deblurring methods, video deblurring benefits from adjacent frames.
Su et al.~\cite{su2017deep} proposed to align the neighbor frames by homograph or optical flow and use a stack of warped frames as input to transfer the image deblurring method to video task. Instead of directly aligning the stack of frames, ~\cite{wang2019edvr, zhou2019stfan} 
tried to align the frames in feature space utilizing attention mechanisms. Liu et al.~\cite{LiuJPSG20} relied on optical flow for a self-supervised pipeline with additional camera parameters.

\textbf{Blur generation.} 
Obtaining corresponding pairs of sharp and blurred images is of great value for deep learning-based deblurring. The most common approach to blur simulation is to use the average of stacks of sharp frames from video shot by high frame-rate camera~\cite{Nah_2019_CVPR_Workshops_REDS,nah2017deep,sun2015learning}. Based on this approach, Brooks et al.~\cite{brooks2019learning} proposed a method to apply frame interpolation techniques on a pair of sharp images to produce an abundance of data. These approaches required external reference other than a single image, and the blur synthesized from specific sharp images was fixed. An alternative approach is to convolve the images by ``camera shake” kernels. Sun et al.~\cite{sun2015learning} used one out of 73 possible linear motion kernels, and Xu et al.~\cite{xu2014deep} also used linear motion kernels. Recently, Kupyn et al.~\cite{KupynBMMM18} applied the Markov process to generate trajectories, which were interpolated to form kernels.

\textbf{Internal learning.}
Recently, researchers are interested in internal learning instead of learning on a big dataset. Ulyanov et al.~\cite{DIP} showed that the structure of a generator network is capable of capturing image statistics prior to any learning. Shocher et al.~\cite{zeroshot} exploited the internal information inside a single image for super-resolution. 

Inspired by these works, the concept of internal learning is widely applicable to many tasks. ~\cite{shaham2019singan,shocher2019ingan} proposed a GAN-based framework to capture the internal distribution of patches within a single image, which could be used in a wide range of image manipulation tasks. To complete blind image deblurring, ~\cite{asim2019blind,ren2019neural} modified the vanilla DIP~\cite{DIP} work to capture the statics of the clear image and the blur kernel at the same time. Furthermore, for the tasks of video, Zhang et al.~\cite{zhang2019internal} proposed an approach built upon DIP~\cite{DIP} and optical flow~\cite{sun2018pwc} to conduct video inpainting.

\section{Method}
\subsection{Overview}
In the traditional learning-based video deblurring pipeline, the model is first trained on a large-scale dataset for a long time and converges to a single pre-trained model. Then the pre-trained model is used on test data to infer the result. However, this kind of pipeline usually involves a very complex model and cannot solve the domain gap problem between training dataset and test data. For example, a pre-trained model trained in scenario A with high performance usually performs worse on a different scenario B.
Kupyn et al.~\cite{kupyn2019deblurgan} reduce the domain gap by estimating the residual between blur and clear image. 
Although they successfully improved the performance on test data with a different scenario, it still brings some artifacts.

As shown in Figure~\ref{fig:pipeline}, unlike the traditional pipeline, we propose a simple but efficient pipeline for video deblurring that addresses the domain gap problem. Some recent approaches, such as SinGAN~\cite{shaham2019singan}  and InGAN~\cite{shocher2019ingan} demonstrate that the information in a single image is enough to train a model. 
Since frames in a short period contain similar texture information, we can use those sharp frames to refine adjacent blurry frames. In this way, we eliminate the use of large training datasets and directly train the network solely on test data. We also noticed that Cho et al.~\cite{cho2012video} have a similar observation to us. They use classical methods rather than a learning-based method to refine blurry frames. As shown in Figure~\ref{fig:TOG}, our method can generate images with higher quality and fewer artifacts.

Figure~\ref{fig:pipeline} illustrates our proposed pipeline. For a test video with motion blur, we firstly select one sharp frame among every 20 frames based on the variance of the Laplacian map. Since we usually have less than ten selected frames, we employ data augmentation on these frames by randomly cropping the chosen frames into $256\times 256$ patches. We did not find rotation or flipping useful in this task because these data augmentation methods create non-realistic texture information that prevents our model from fitting to the selected frames. Taking those patches as ground truth, we randomly select generated blur kernels and make blurry patches fed into the network as input. After fitting, this on-the-fly model will be used to infer each frame of this video.

\subsection{Implementation}
\subsubsection{Frame selection}
The variance of the image Laplacian can be considered as the measurement of the degree of blurriness since it is proportional to the sharpness of an image~\cite{laplacian}. Given an image $I$, the variance of image Laplacian is
\begin{equation}
    M_{VL} = \sum_{(i,j)\in I}{(\Delta I(i,j)-\overline{\Delta I})^2},
\end{equation}
where $\Delta I$ is the image Laplacian obtained by convolving $I$ with the Laplacian mask and $\overline{\Delta I}$ is the mean value of image Laplacian in $I$. 

There is a trade-off between texture information and the complexity of convergence. Increasing the number of chosen frames will bring more texture information but increase the time and difficulty for the network convergence. Thus, we empirically choose the frame with the highest $M_{VL}$ among every 20 frames to ensure the selected frames cover all texture information in the video.

\subsubsection{Blur generation}
Since sharp images alone are not enough to train a deblurring network, we need to obtain pairs of corresponding sharp and blurred images with a suitable blur generation strategy. There are three major factors we should take into consideration about blur generation.

First, we notice that the motion blur is mainly caused by camera motion in most cases. Though the camera trajectory may be complex for a long period, the blur kernel is approximately linear in a short period. We represent the local blur kernel $M$ of size $p \times p$ by a motion vector $m = (l, o)$, which characterizes the length and orientation of the motion field~\cite{sun2015learning} ($l \in (0,p)$ and $ o \in [0,180^\circ)$). Then the kernel is generated by applying sub-pixel interpolation to the motion vector.

\begin{figure}[t]
\centering
\begin{tabular}{c@{\hspace{0mm}}c@{\hspace{0.1mm}}c@{\hspace{0mm}}c@{\hspace{0.1mm}}c@{\hspace{0mm}}c}
\includegraphics[width=0.16\columnwidth]{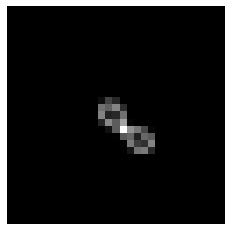}&
\includegraphics[width=0.16\columnwidth]{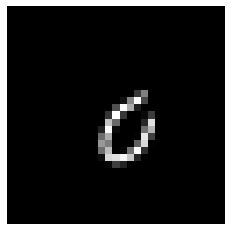} &
\includegraphics[width=0.16\columnwidth]{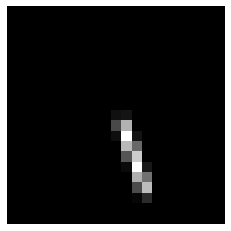} &
\includegraphics[width=0.16\columnwidth]{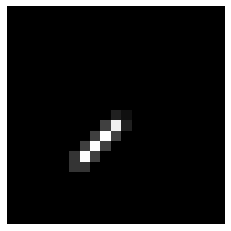}&
\includegraphics[width=0.16\columnwidth]{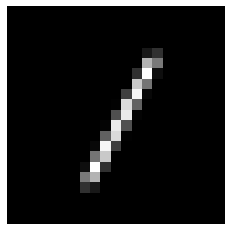}&
\includegraphics[width=0.16\columnwidth]{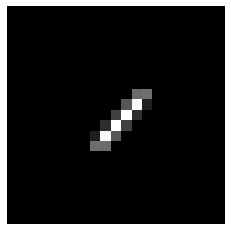}\\
\multicolumn{2}{c}{Simulated~\cite{KupynBMMM18}} & \multicolumn{2}{c}{Asymmetric} & \multicolumn{2}{c}{Ours}\\
\end{tabular}
\caption{Visualization of three kinds of blur kernels. Our blur kernel is symmetric and linear, which leads to better video deblurred results.}
\label{fig:kernel}
\vspace{-1em}
\end{figure}

Second, a video frame with motion blur could be considered as the average of a period of time $t$. We can assume the ground-truth sharp frame is at time $\frac{t}{2}$ for simplicity. Thus we propose to use symmetry blur kernels, which means the center of $m$ is always at the center of the blur kernel $M$. In Section~\ref{section:exp}, we empirically show using symmetry blur kernels improves both spatial and temporal stability, compared to simulated blur kernels~\cite{KupynBMMM18} and asymmetric kernels. 
Figure~\ref{fig:kernel} visualizes the different kinds of blur kernels.

Third, we consider improving blur generation in the nearly raw data space, rather than the RGB color space. In reality, the motion blur already occurs in raw data before the image signal processing pipeline generates an RGB image. In this way, synthesizing blur in the raw data space is preferred. To simulate the intensity in raw data, we reverse the gamma correction operation. After reversing the gamma correction, the pixel values are roughly linear. 

In the end, after selecting sharp patches from patch generation, we perform reversed gamma correction and apply blur kernels to generate patches with motion blur. To reconstruct these images in the RGB space, we apply the gamma correction to them.

\subsubsection{Network}
 
In the traditional video deblurring pipeline, the model is trained on an extensive collection of training data with diverge scenarios. To capture different cases, the state-of-the-art models are usually deep with complex components, i.e., deformable convolution~\cite{dai2017deformable} and spatial attention, and hence require high computing power and a long period for training. 
For our pipeline, we found that such deep networks were unnecessary. The performance might even reduce because the small number of training images is not enough for these deep networks.
Instead, we find that a simple U-Net~\cite{Unet} is enough to fit our training data well. With our pipeline, the network can aggregate the information in nonlocal frames instead of several neighboring frames, without the need of optical flow.

\subsubsection{Loss function}
Rather than using the common pixel-space loss, e.g., the most straightforward $L_2$ distance, which tends to make the result smooth, we adopt perceptual loss to make a sharp result and preserve more high-frequency information.

Given a trained visual perception network $\Phi$ (we use VGG-19~\cite{SimonyanZ14a}), we define a collection of layers $\Phi$ as $\{ \Phi_{l} \}$. For a training pair $\left( I, L\right) $, where $L$ is the ground truth image patch and $I$ is the input image patch, our perceptual loss is
\begin{equation}
    \mathcal{L}_{P} = \sum_{l}\lambda_{l}\Arrowvert\Phi_{l}(L)-\Phi_{l}(G(I)) \Arrowvert_{1}.
\end{equation}
Here $G$ is the U-Net in our pipeline. The hyperparameters $\{\lambda_{l}\}$ balance the contribution of each layer $l$ to the loss.

Different from the traditional pipeline, which refers to sharp ground truth images, our method tries to absorb the information of relatively sharp and clear images contained in the test video. Thus, we apply a reweighting strategy to force the network to pay more attention to images of higher sharpness. Note that the variance of image Laplacian $M_{VL}$ used in our pipeline indicates the sharpness of a patch. Thus we normalize the variance by a constant $N$ and multiply it with the original loss function as a reweighted loss function:
\begin{equation}
    \mathcal{L}_{reweighted} = \frac{M_{VL}}{N} \mathcal{L}_{P}.
\end{equation}
Moreover, to increase the network's focus on the textural information and further improve the result, we add a Markovian discriminator $D$ that has five hidden layers with WGAN-GP loss~\cite{wgan,LiW16,shaham2019singan} to our pipeline. 

\subsection{Acceleration}
Our pipeline exploits the internal information of sharp frames in the test video. However, it usually takes hours to fit a model to these sharp frames for video deblurring due to the complexity and a massive number of possible blur kernels. In our experiments, we notice that the network takes a long period to adapt itself to generate expected sharp images at the beginning of the training. We first used a simple pretraining strategy and found the pretrained model would be easily stuck at a local minimum. 

Inspired by ~\cite{finn2017model,park2020fast}, we propose to initialize the network via meta-learning, which enables rapid fitting on any given video at test time. We find that model-agnostic meta-learning (MAML)~\cite{finn2017model} can make the fine-tuning process easier as it provides our model a proper initialization.
MAML plays a crucial role in meta-learning due to its high compatibility with various models and tasks by using gradient descent. In our accelerated pipeline, we first initialize our network with ten external test videos (without ground truth) via MAML to lead the model to learn the internal information. At the inference stage, we fine-tune the meta-learned parameters using the given test video in about 5 minutes. The details of the MAML algorithm for deblurring are shown in Algorithm~\ref{algorithm:train}, where $\{I_i\}$ denotes the sharp frames sampled from a small set of test videos, $\{GT_i\}$ denotes the sharp patches cropped from $\{I_i\}$ ,$\{Blur_i\}$ and $\{Blur^*_i\}$ denote blurry patches generated based on $\{GT_i\}$ using two different blur kernels and $g_{\theta}$ denotes our network initialized with the parameter $\theta$.

\begin{algorithm}[t]
\caption{ MAML algorithm for deblurring} 
\hspace*{0.02in} {\bf Require:} 
$p(I)$: Distribution over images\\
\hspace*{0.02in} {\bf Require:} 
$D$: a list of blur kernels \\
\hspace*{0.02in} {\bf Require:} 
$\alpha,\beta$: Hyper-parameters (step size)

\begin{algorithmic}[1]
\State Initialize $\theta$ 
\While{not converged} 
    \State Randomly sample a batch of images $\{I_i\} \sim p(I)$ and two blur kernels from $D$
    \State Random crop for image patches $\{GT_i\}$ from $\{I_i\}$
    \State Generate $\{Blur_i\}$, $\{Blur^*_i\}$ from $\{GT_i\}$
    \For{each $i$}
    \State
    Evaluate $\nabla_{\theta}\mathcal{L}(g_{\theta}(Blur_i),GT_i)$ using $\mathcal{L}$
    \State
    Compute adapted parameters
    $\theta_i \leftarrow \theta - \alpha \nabla_{\theta}\mathcal{L}(g_{\theta}(Blur_i),GT_i)$
    \EndFor
    \State Update $\theta \leftarrow \theta - \beta \nabla_{\theta} \sum_{i} \mathcal{L}(g_{\theta_i}(Blur^*_i),GT_i)$
\EndWhile
\end{algorithmic}
\label{algorithm:train}
\end{algorithm}

\begin{figure*}[t]
\centering
\begin{tabular}{@{}c@{\hspace{1mm}}c@{\hspace{1mm}}c@{\hspace{1mm}}c@{}}
&\includegraphics[width=0.32\linewidth]{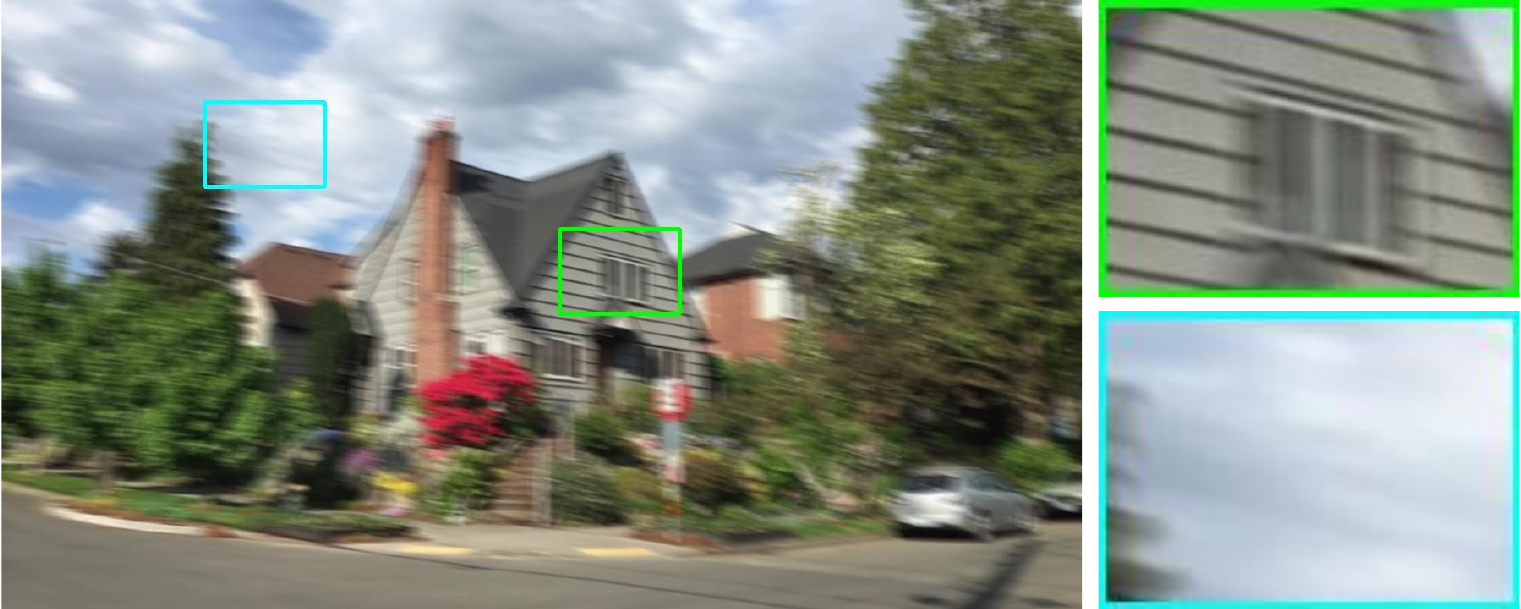}&
\includegraphics[width=0.32\linewidth]{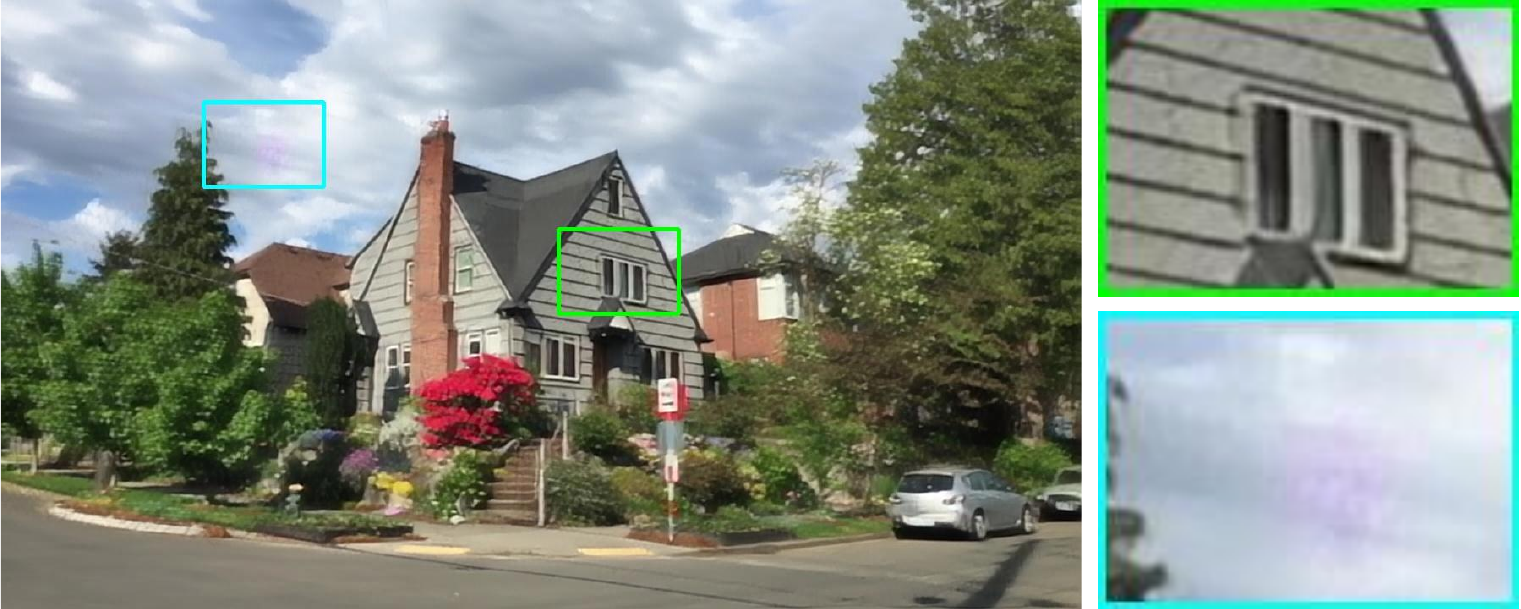}&
\includegraphics[width=0.32\linewidth]{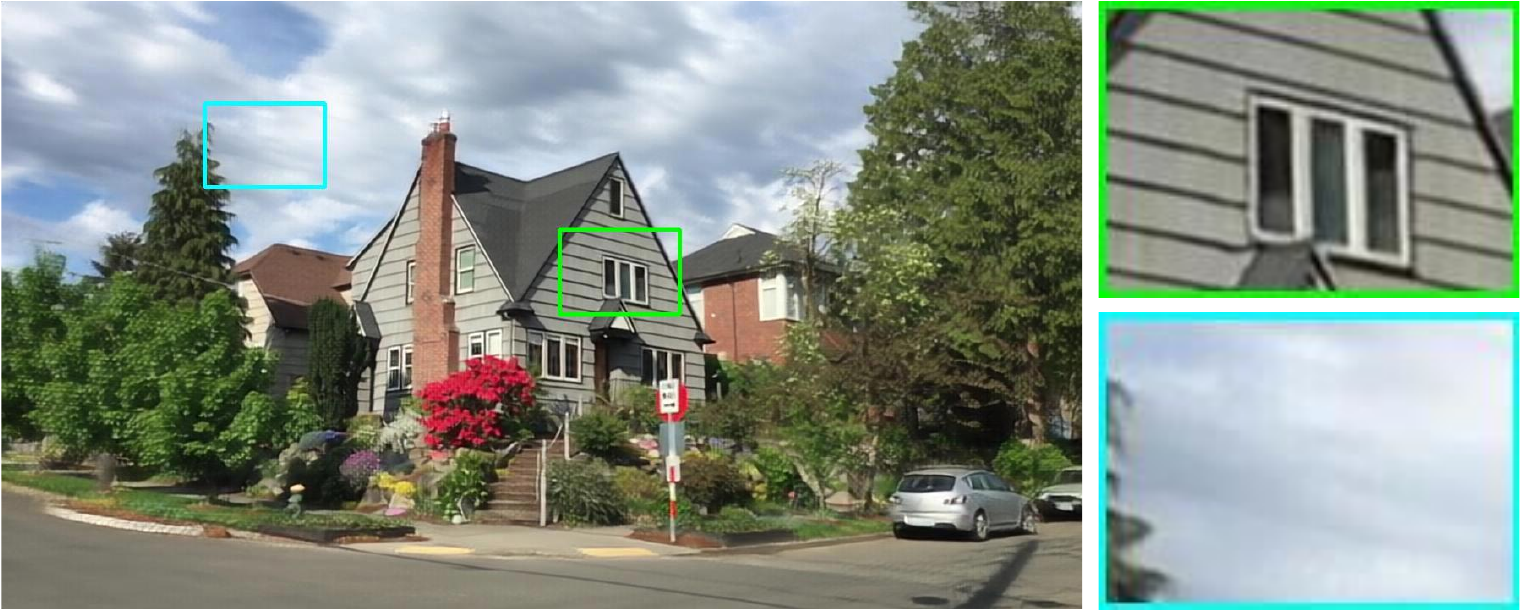}\\
&Input & DeblurGAN-v2~\cite{kupyn2019deblurgan} & Ours  \\
&{PSNR/SSIM} & 25.02/0.902 & 24.34/0.889 \\
\end{tabular}
\caption{The results on an example in the synthetic dataset. Note that our deblurred result is sharper and clearer than the result by DeblurGAN-v2 \cite{kupyn2019deblurgan}, but the PSNR and SSIM do not reflect that our result is better.}
\label{fig:SSIM}
\end{figure*}

\section{Experiments}
\label{section:exp}

\subsection{Dataset}

In prior video deblurring work, synthetic datasets are often used for evaluation because there are no ground-truth sharp videos for real-world blurry videos. However, we should better evaluate different methods on real-world blurry videos rather than synthetic data. We find that model training on synthetic data does not necessarily perform well on real-world data. Therefore, we collected a real-world video dataset, denoted as \textit{Real-World}, with 70 videos with motion blur for qualitative and quantitative evaluation. The videos are captured by shaking, walking, or running in diverse indoor and outdoor environments. Each video contains 80-160 frames, shot with hand-held devices, including iPhone 8 Plus, iPhone 11 Pro Max, and Huawei Mate 20. For the self-driving car scenario, we use a popular dataset \textit{BDD100K}~\cite{BDD100K}, consisting of video sequences of driving experience across many different times in the day, weather conditions, and driving scenarios. Among thousands of videos, we randomly select 20 videos with motion blur.

Although there are no ground-truth sharp videos in \textit{BDD100K} and \textit{Real-World} dataset, we can still conduct a user study to compare different methods quantitatively. For the temporal consistency evaluation, we use two synthetic datasets: \textit{GOPRO} dataset~\cite{su2017deep} and \textit{REDS} dataset~\cite{Nah_2019_CVPR_Workshops_REDS}.

\subsection{User study}

\begin{table}[t]
\centering
\caption{User study results of pairwise comparisons between several baselines and our pipeline.
Each cell lists the fraction of pairwise comparisons in which images synthesized by our approach were rated sharper and clearer than images synthesized by the corresponding baseline.}
\begin{tabular}{l@{\hspace{2em}}c@{\hspace{2em}}c}
\toprule
Comparison & BDD100K & Real-World \\
\midrule  
\textit{Image-based methods} \\
Ours $>$ SRN~\cite{tao2018srndeblur}& 69.6\% &  85.7\% \\
Ours $>$ DeblurGAN-v2~\cite{kupyn2019deblurgan}& 66.0\% &  80.1\% \\
Ours $>$ SelfDeblur~\cite{ren2019neural} & 90.7\% & 93.4\%\\
\midrule  
\textit{Video-based methods} \\
Ours $>$ EDVR~\cite{wang2019edvr} & 74.6\% & 85.5\% \\
Ours $>$ DMPHN~\cite{Zhang_2019_CVPR} & 71.1\% & 94.2\% \\
Ours $>$ STFAN~\cite{zhou2019stfan} & 66.7\% & 79.6\% \\ 
\bottomrule
\end{tabular}
\label{tbl:Metric}
\end{table}

\begin{table}[t]
\centering
\caption{User study results on the Real-World dataset between several baselines and our MAML pipeline. Ours-MAML is preferred by most of the users.}
\begin{tabular}{l@{\hspace{5mm}}c}
\toprule
Comparison & Preference rate \\
\midrule  
\textit{Image-based methods} \\
Ours-MAML $>$ SRN~\cite{tao2018srndeblur}~\cite{Zhang_2019_CVPR} & 83.7\% \\
Ours-MAML $>$ DeblurGAN-v2~\cite{kupyn2019deblurgan} & 79.0\% \\
Ours-MAML $>$ SelfDeblur~\cite{ren2019neural} & 91.2\% \\
\midrule  
\textit{Video-based methods} \\
Ours-MAML $>$ EDVR~\cite{wang2019edvr} & 84.7\%\\
Ours-MAML $>$ DMPHN~\cite{Zhang_2019_CVPR} & 92.7\% \\
Ours-MAML $>$ STFAN~\cite{zhou2019stfan} & 79.4\% \\
\bottomrule
\end{tabular}
\label{tbl:MAML}
\vspace{-1em}
\end{table}


\begin{table}[t]
\centering
\caption{Performance comparison in temporal consistency metric (lower is better). Our method with symmetric kernels achieves the best temporal consistency among all the baselines. }
\begin{tabular}{l@{\hspace{5mm}}c}
\toprule
Method & $E_{warp}\downarrow$ \\
\midrule  
DMPHN~\cite{Zhang_2019_CVPR} & 0.3155 \\
EDVR~\cite{wang2019edvr} & 0.2874 \\
STFAN~\cite{zhou2019stfan} & 0.2739 \\
DeblurGAN-v2~\cite{kupyn2019deblurgan} & 0.2689 \\
\cdashlinelr{0-1}
Ours (with asymmetric kernels) & {0.3274} \\
Ours (with simulated kernels) &  {0.3928} \\
Ours & \textbf{0.2599} \\
\bottomrule
\end{tabular}
\label{tbl:Temporal}
\vspace{-1em}
\end{table}

\begin{figure*}[t]
\centering
\begin{tabular}{@{}c@{\hspace{1mm}}c@{\hspace{1mm}}c@{\hspace{1mm}}c}
\includegraphics[width=0.23\linewidth]{{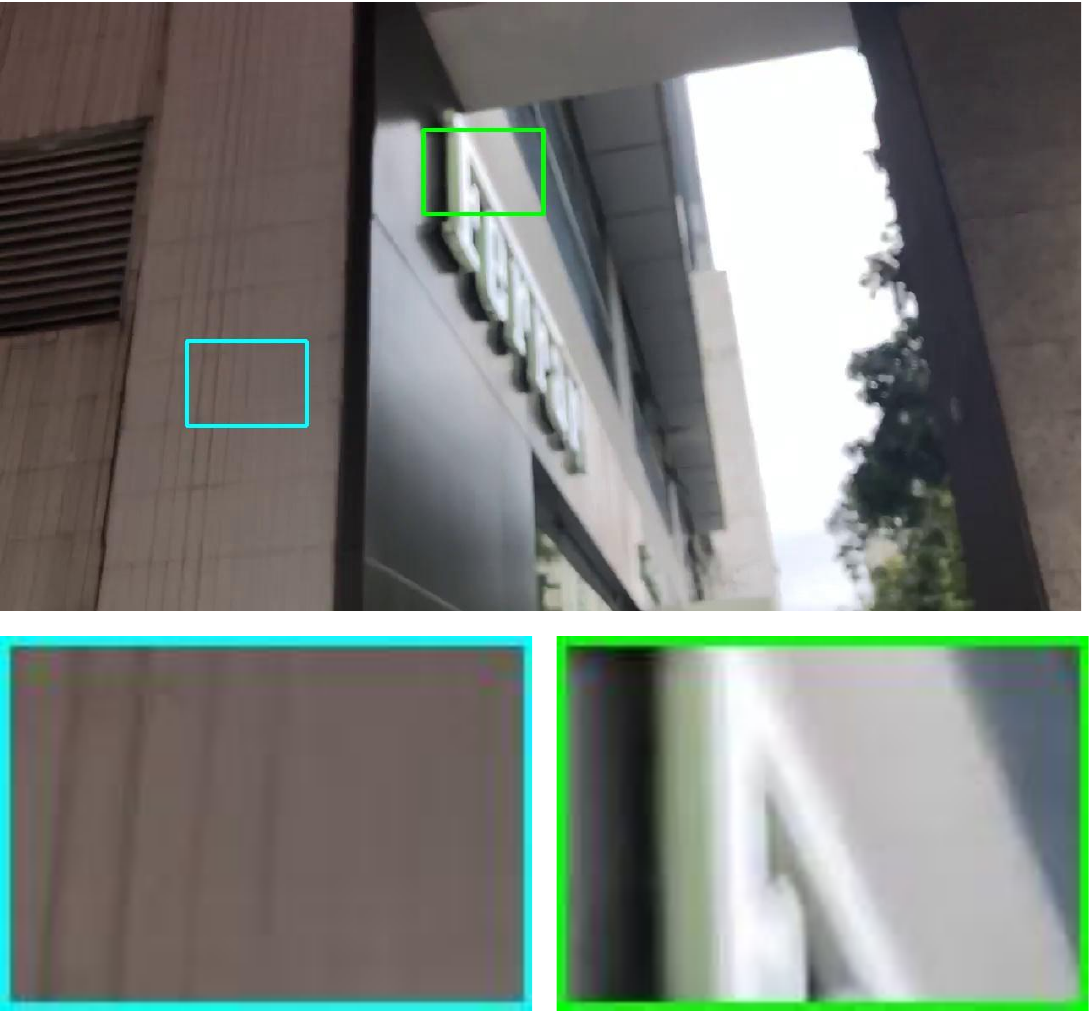}} &
\includegraphics[width=0.23\linewidth]{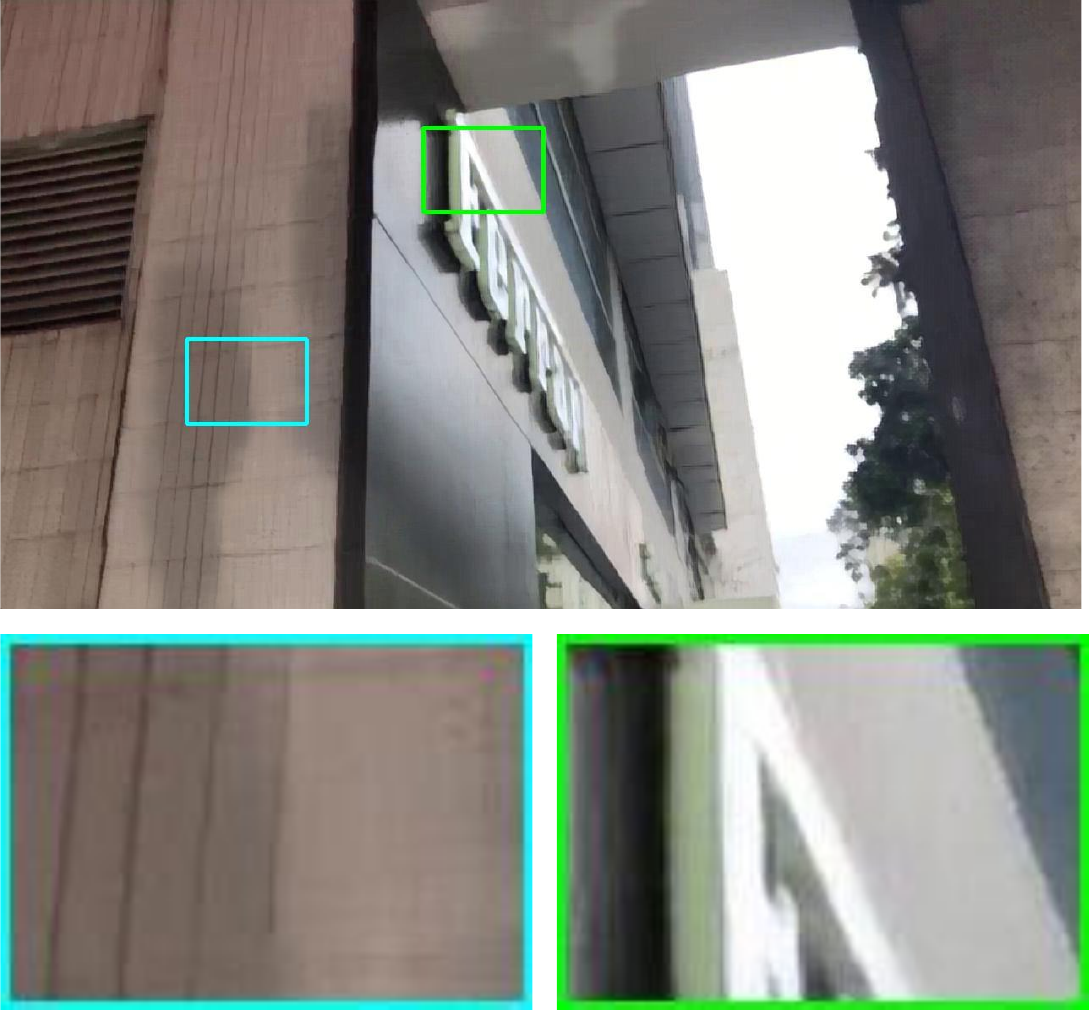} &
\includegraphics[width=0.23\linewidth]{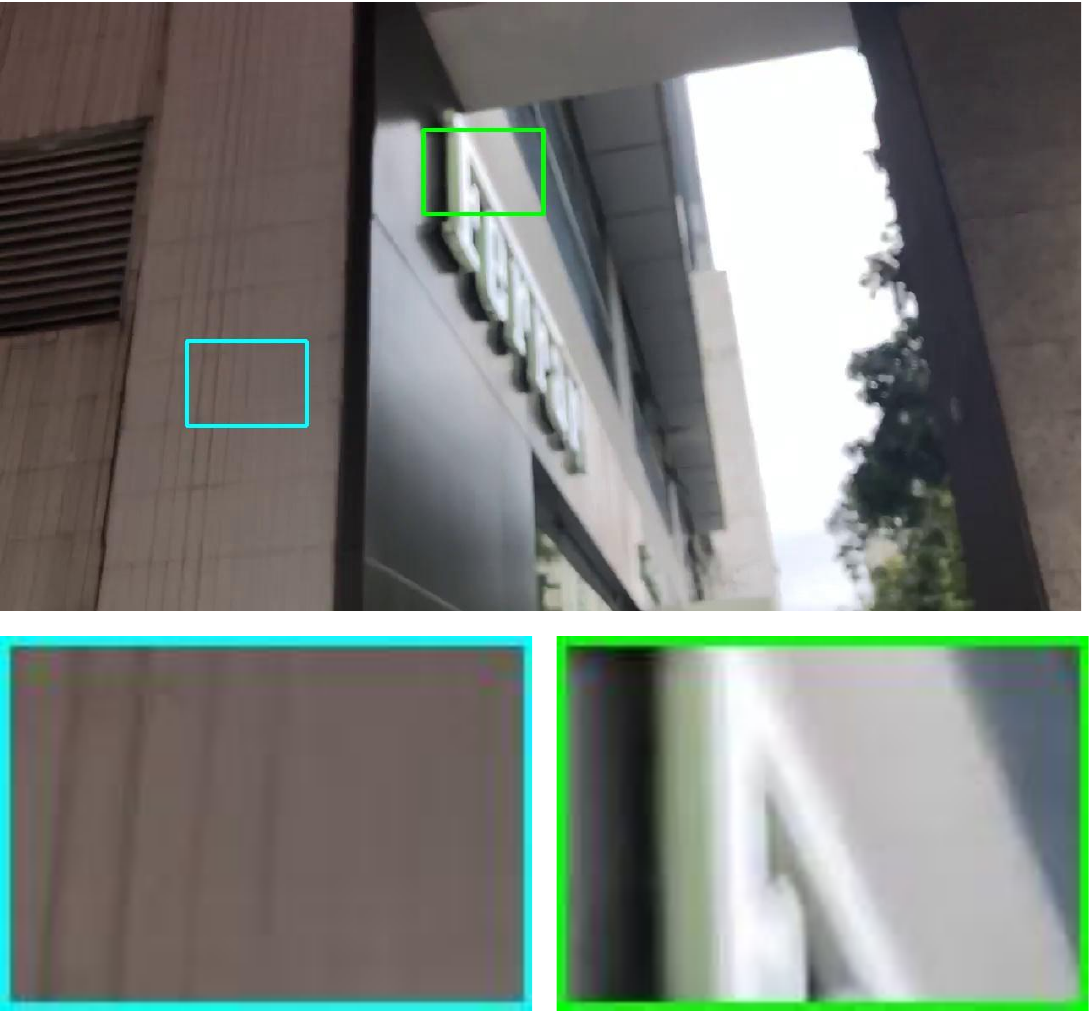} &
\includegraphics[width=0.23\linewidth]{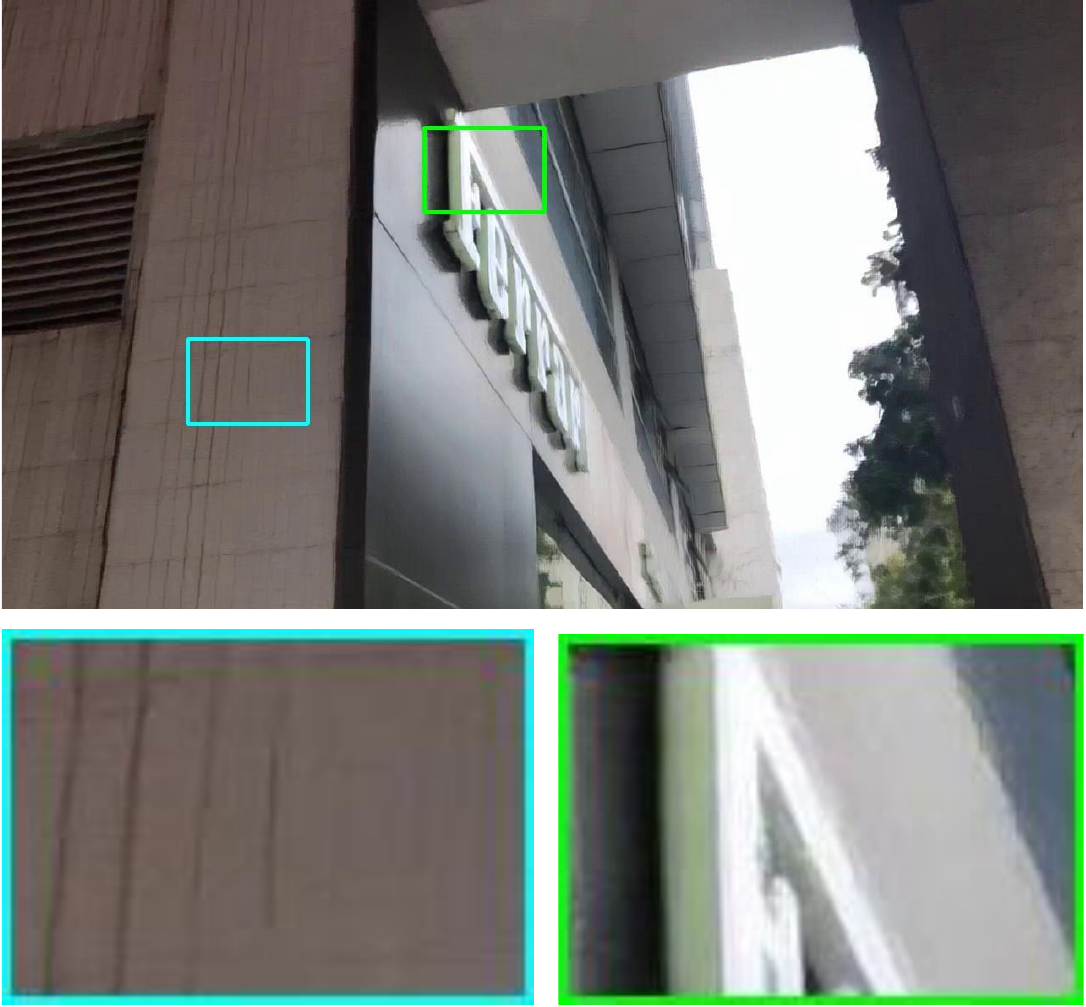}\\
Input& Simulated~\cite{KupynBMMM18} & Asymmetric & Ours\\
\end{tabular}
\caption{The visual ablation study on different kinds of blur kernels. The symmetric blur kernels improve the spatial stability of our pipeline without bringing artifacts.}
\label{fig:comparison_kernel}
\end{figure*}

For image restoration tasks like deblurring and super-resolution, PSNR and SSIM are always used to compare among different methods because these metrics are an approximation to human perception of image quality. However, computing PSNR and SSIM is only possible on synthetic data instead of real-world data for video deblurring. Furthermore, as shown in Figure~\ref{fig:SSIM}, these metrics may not agree with human perception~\cite{zhang2018single}. Our method is able to recover sharp and clear video frames from a synthetic video, while the result by DeblurGAN-v2 is of lower quality. Therefore, we mainly adopt user study as the main evaluation metric for video deblurring. 

To compare our method with state-of-the-art baselines, we conduct an extensive user study on Amazon Mechanical Turk for both fitting-to-test-data pipeline and accelerated pipeline, following the A/B test protocol proposed by Chen and Koltun~\cite{ChenK17}. Among all the prior work, we choose several representative blind image and video deblurring methods for comparisons: SRN (CVPR 2018)~\cite{tao2018srndeblur}, DMPHN (CVPR 2019)~\cite{Zhang_2019_CVPR}, EDVR (CVPRW 2019) ~\cite{wang2019edvr}, DeblurGAN-v2 (ICCV 2019)~\cite{kupyn2019deblurgan}, STFAN (ICCV 2019)~\cite{zhou2019stfan}, and SelfDeblur (CVPR 2020)~\cite{ren2019neural}. For all these baselines, we use their public pre-trained models.

Instead of showing videos, we choose to randomly sample one blurry frame from each video so that participants have more time to focus on the details of the results. The sampled image is cropped into a square of size $720 \times 720$ for better judgment. 
During the user study, each user is presented with an image deblurred by our method and an image deblurred by a baseline simultaneously in a random order in the same row. Then the user needs to choose an image that is sharper and clearer between the two deblurred images. The results are summarized in Table~\ref{tbl:Metric} and Table~\ref{tbl:MAML}.

According to Table~\ref{tbl:Metric}, the rates of Ours are all larger than the baselines on the two datasets, which shows our fitting-to-test-data pipeline is preferred by most users. For our accelerated pipeline, we conduct experiments on the Real-World dataset with general scenarios. To some extent, our accelerated pipeline scarifies the image quality to shorten the running time to about 5 minutes. However, it can still outperform the state-of-the-art methods.

\begin{figure*}[t]
\centering
\begin{tabular}{@{}c@{\hspace{1mm}}c@{\hspace{1mm}}c@{\hspace{1mm}}c@{}}

&\includegraphics[width=0.32\linewidth]{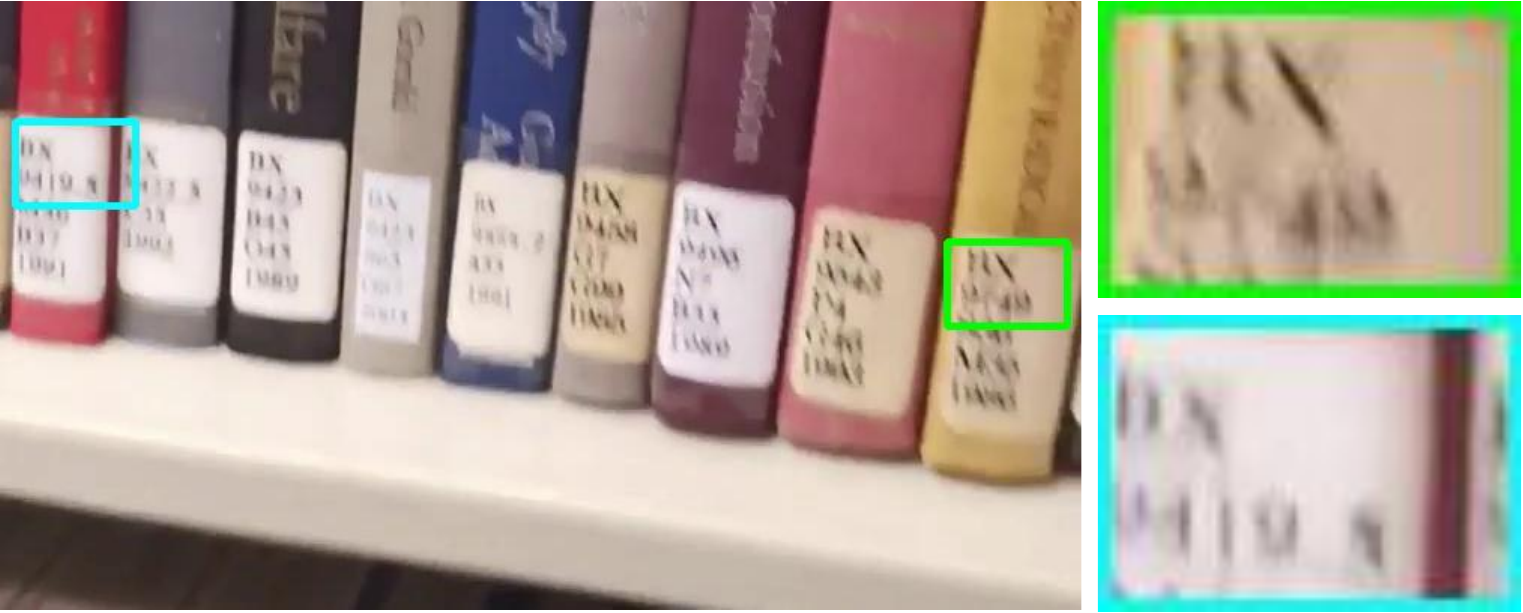}&
\includegraphics[width=0.32\linewidth]{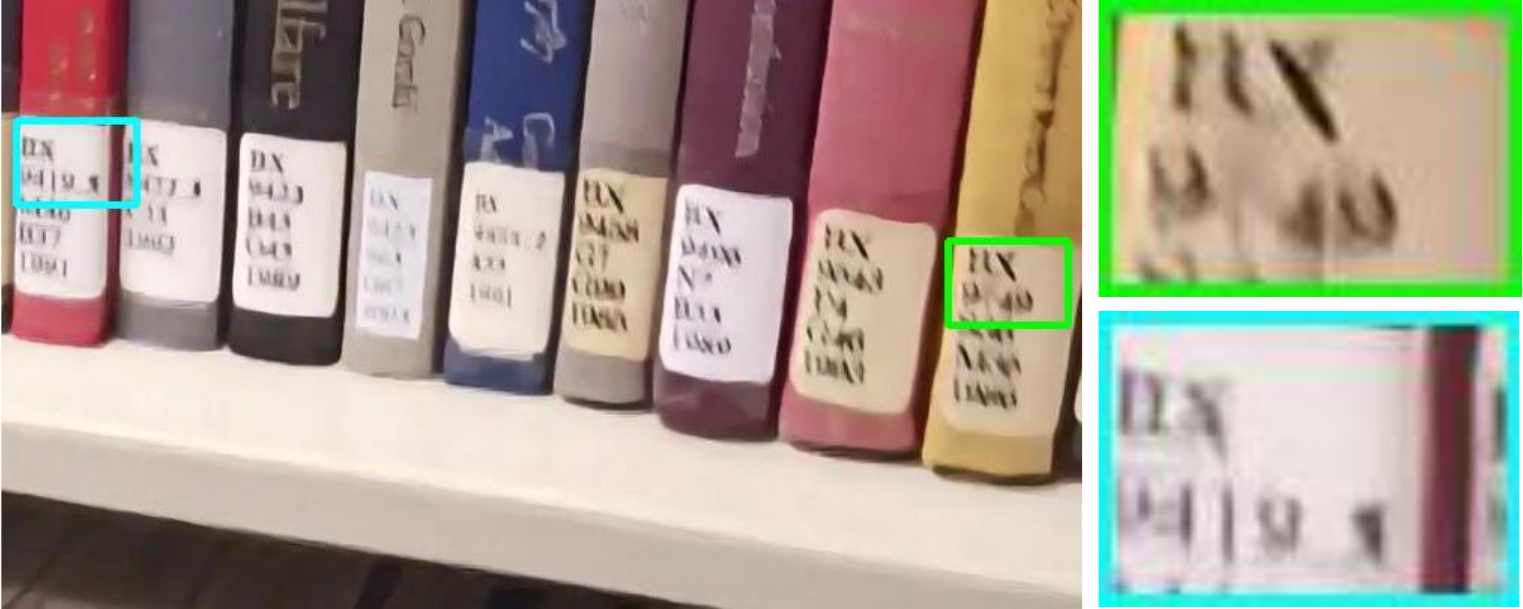}&
\includegraphics[width=0.32\linewidth]{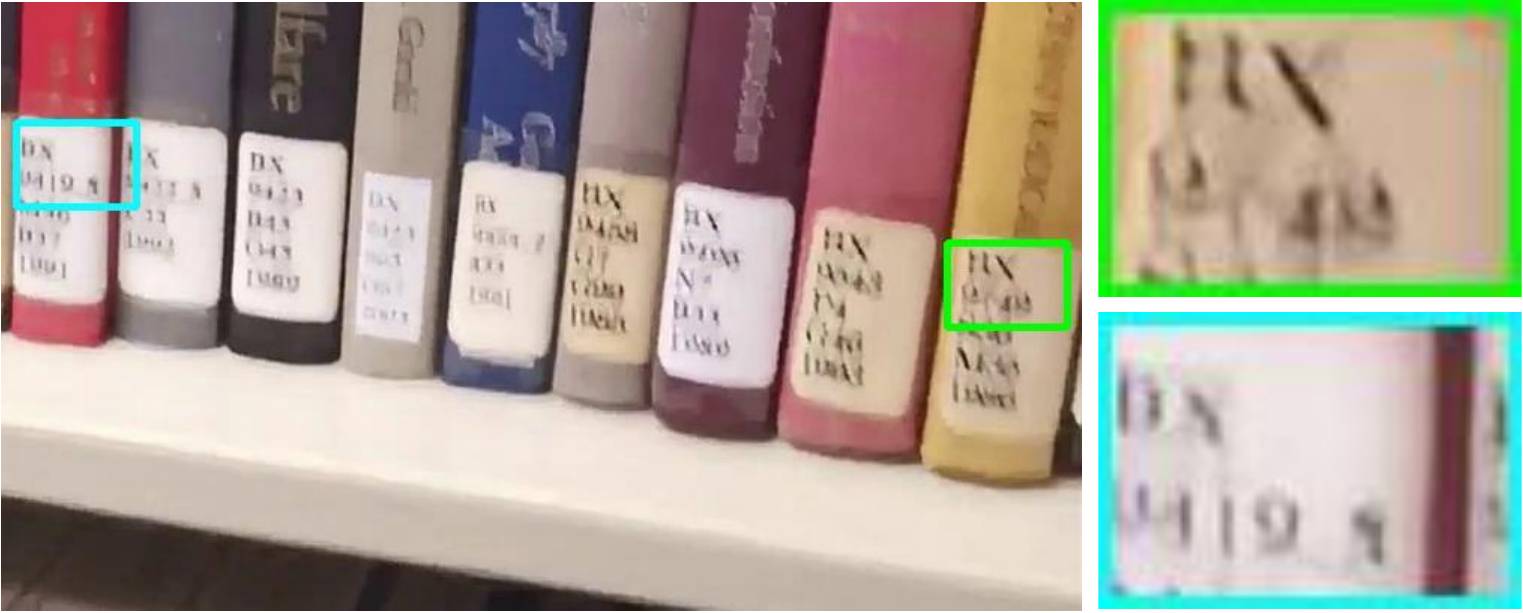}
\\
&Input &EDVR~\cite{wang2019edvr} &DeblurGAN-v2~\cite{kupyn2019deblurgan} \\

&\includegraphics[width=0.32\linewidth]{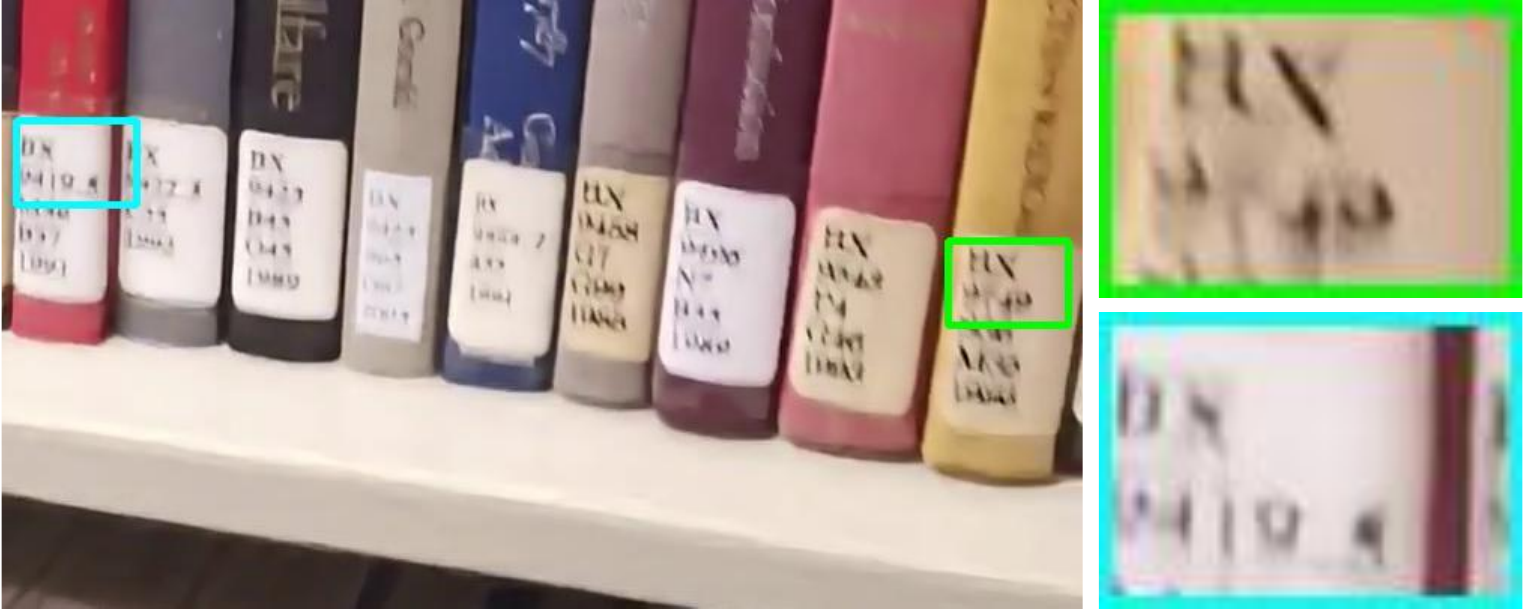}&
\includegraphics[width=0.32\linewidth]{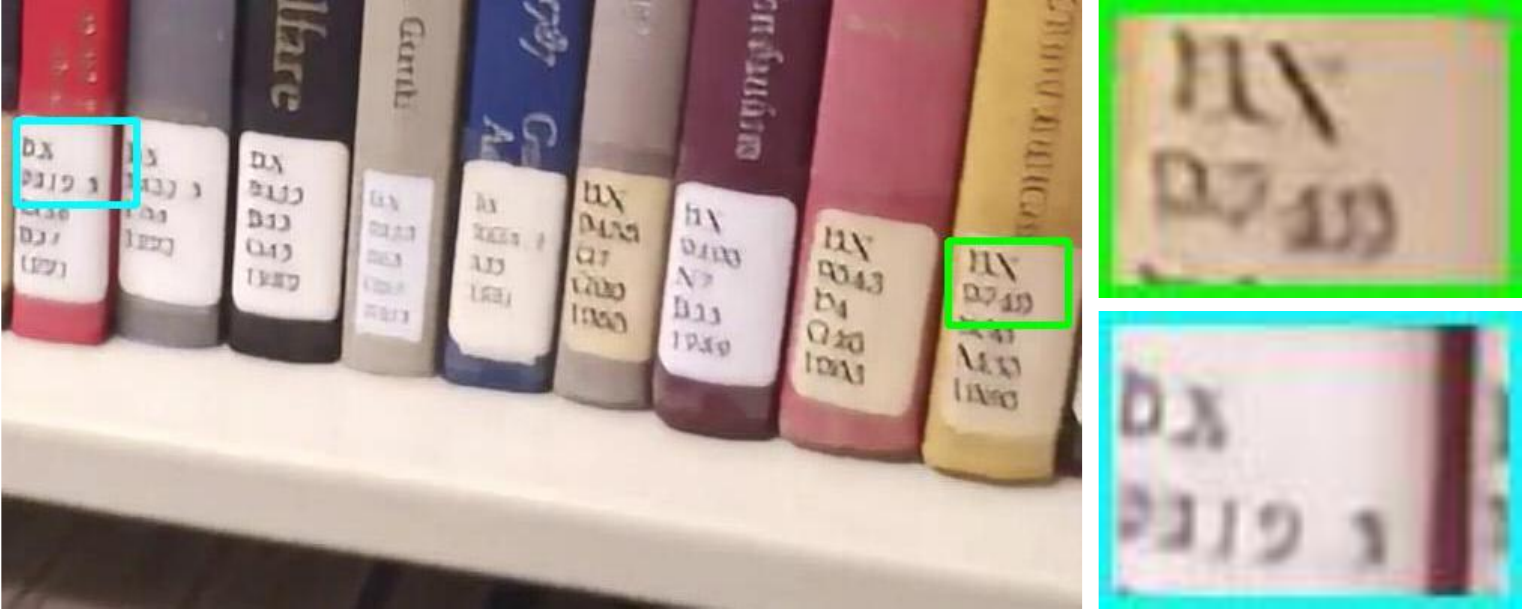}&
\includegraphics[width=0.32\linewidth]{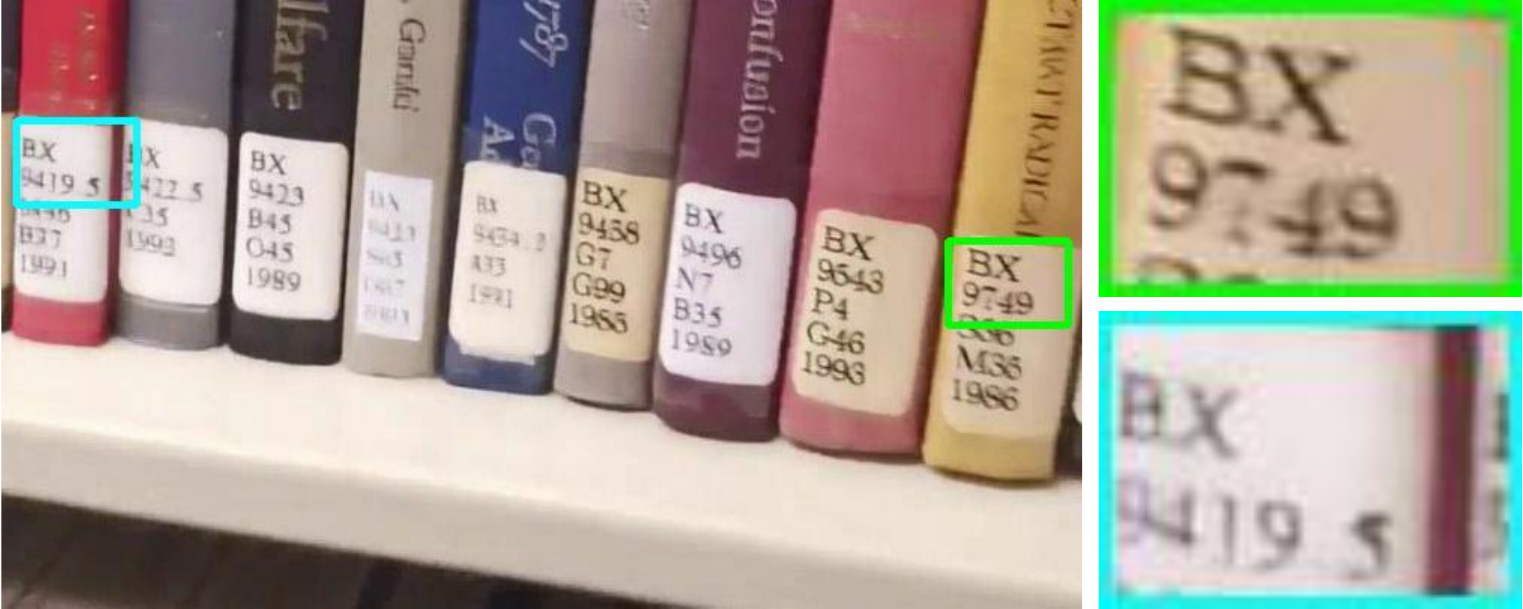}
\\
& STFAN~\cite{zhou2019stfan} &Our MAML & Ours\\

&\includegraphics[width=0.32\linewidth]{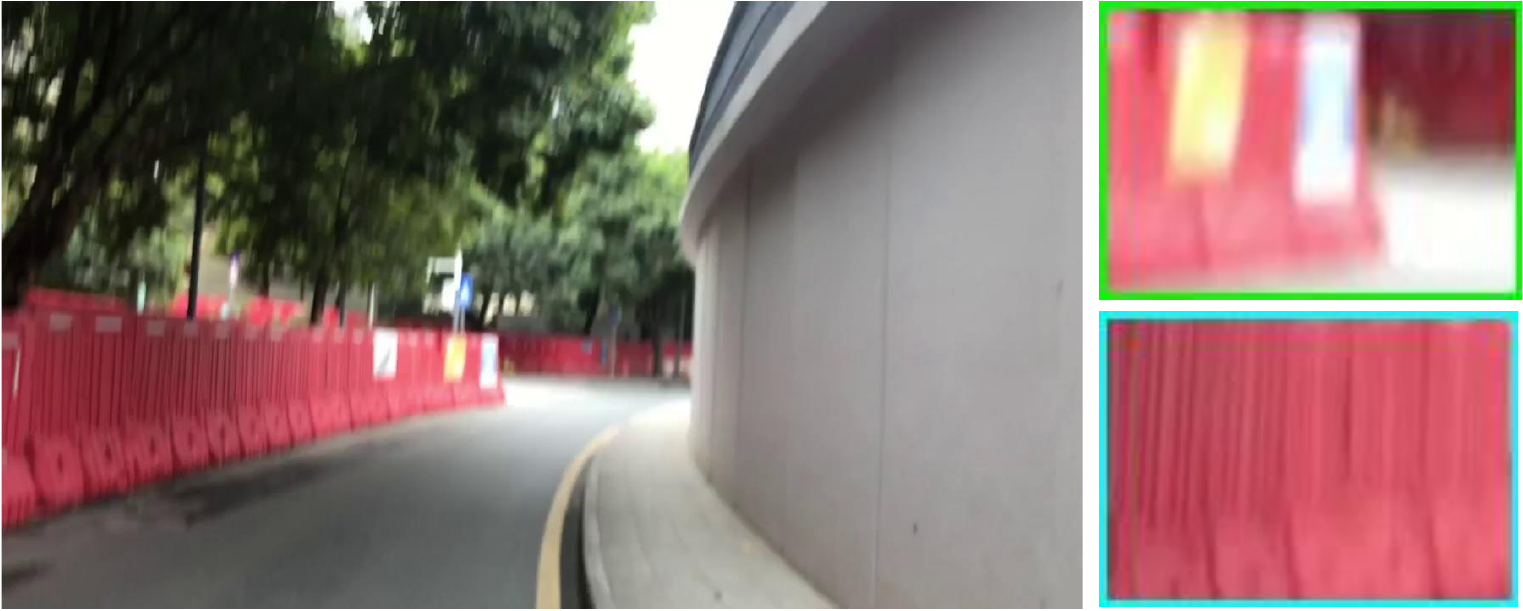}&
\includegraphics[width=0.32\linewidth]{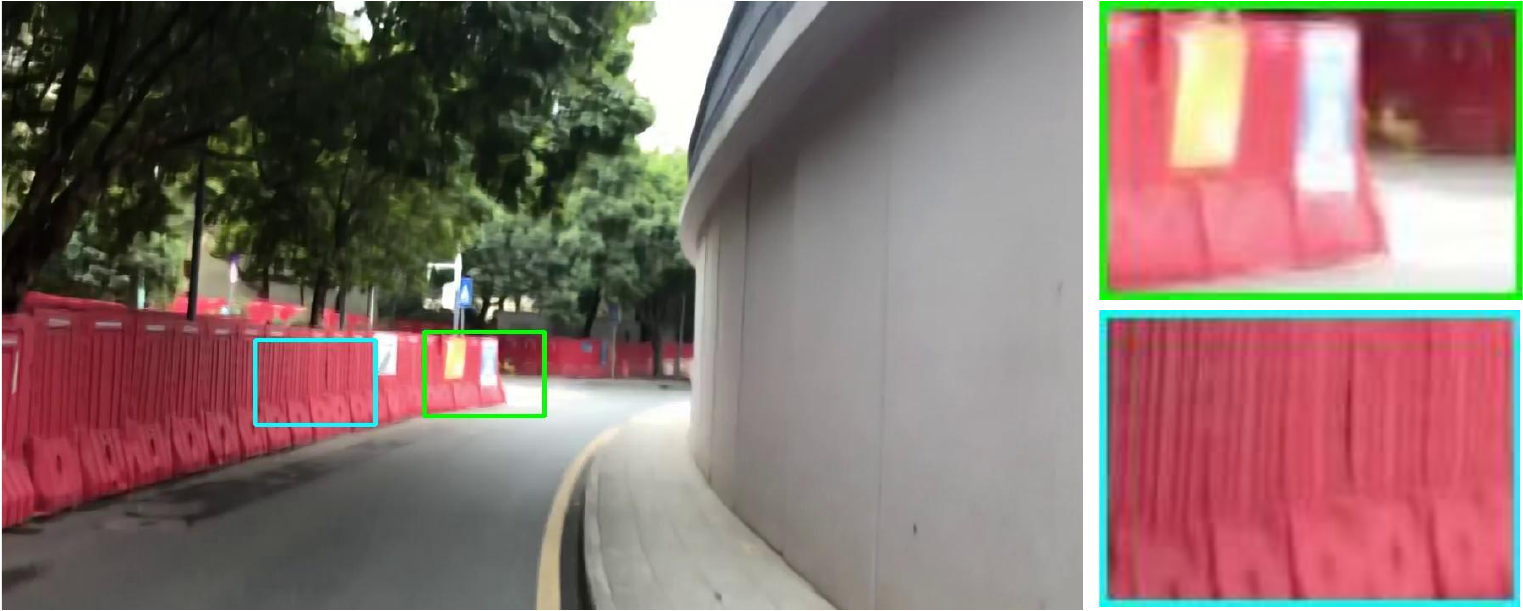}&
\includegraphics[width=0.32\linewidth]{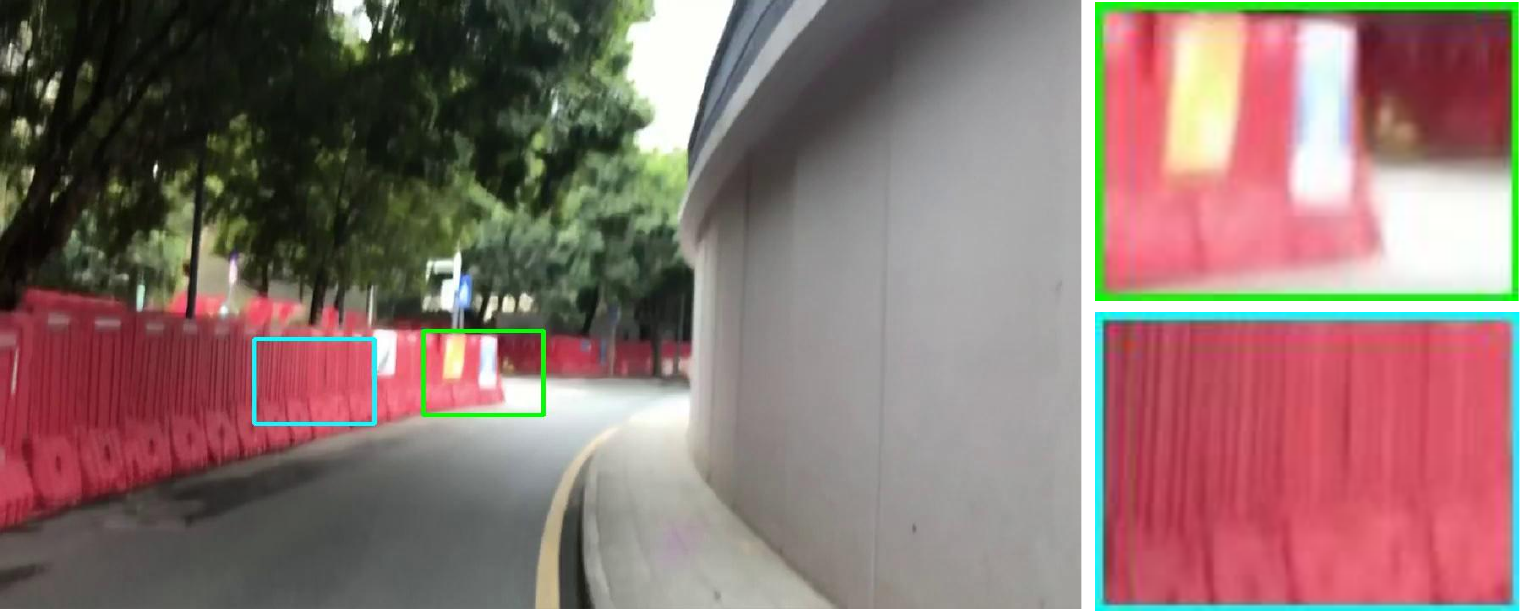}
\\
&Input &EDVR~\cite{wang2019edvr} &DeblurGAN-v2~\cite{kupyn2019deblurgan} \\

&\includegraphics[width=0.32\linewidth]{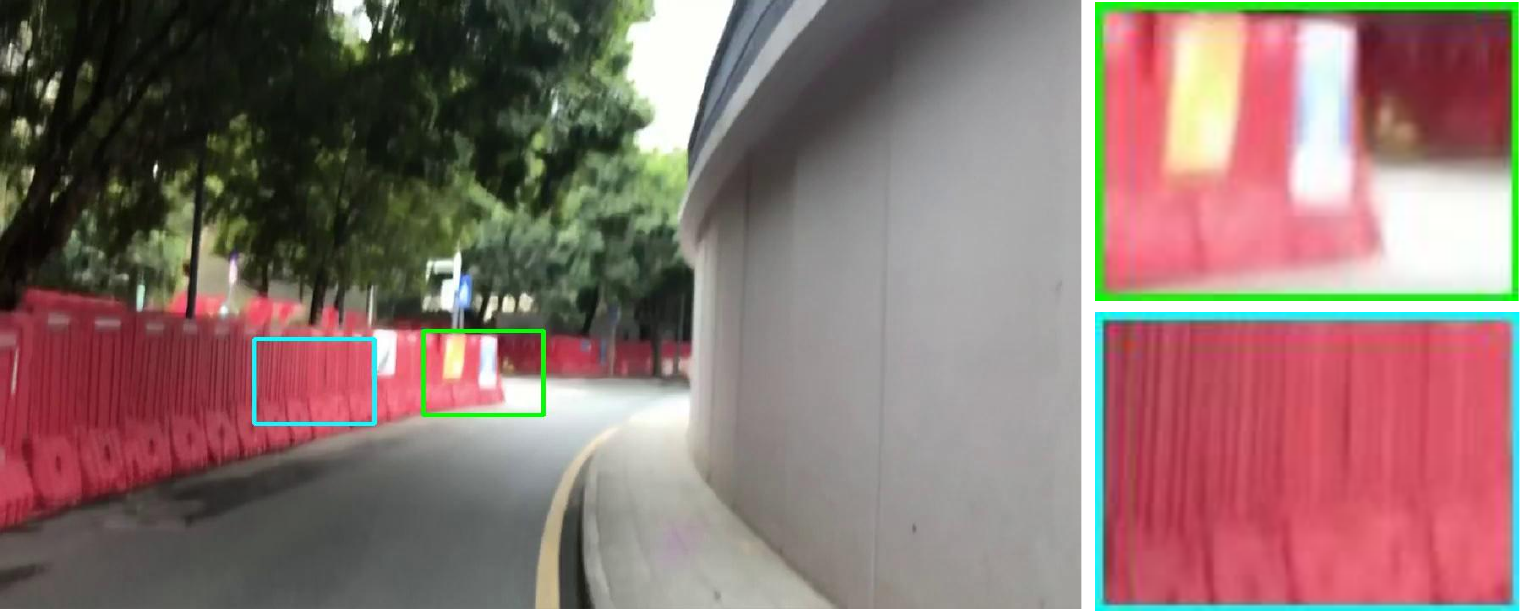}&
\includegraphics[width=0.32\linewidth]{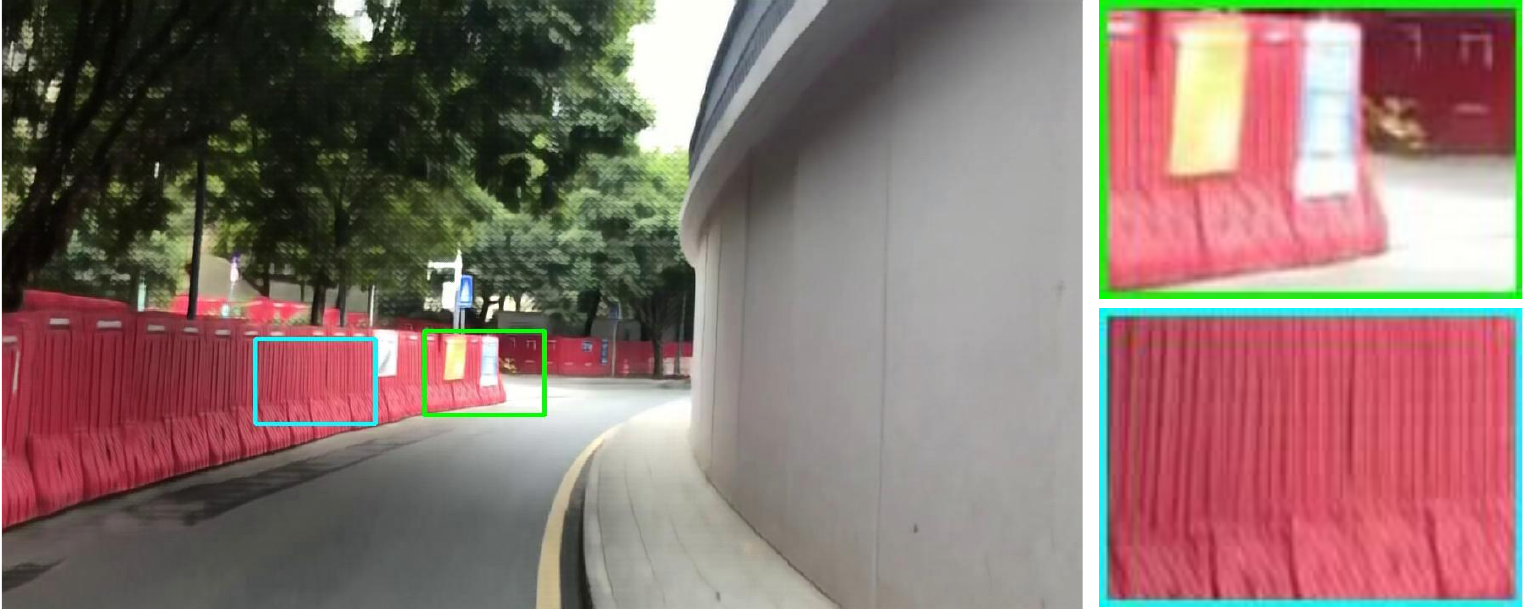}&
\includegraphics[width=0.32\linewidth]{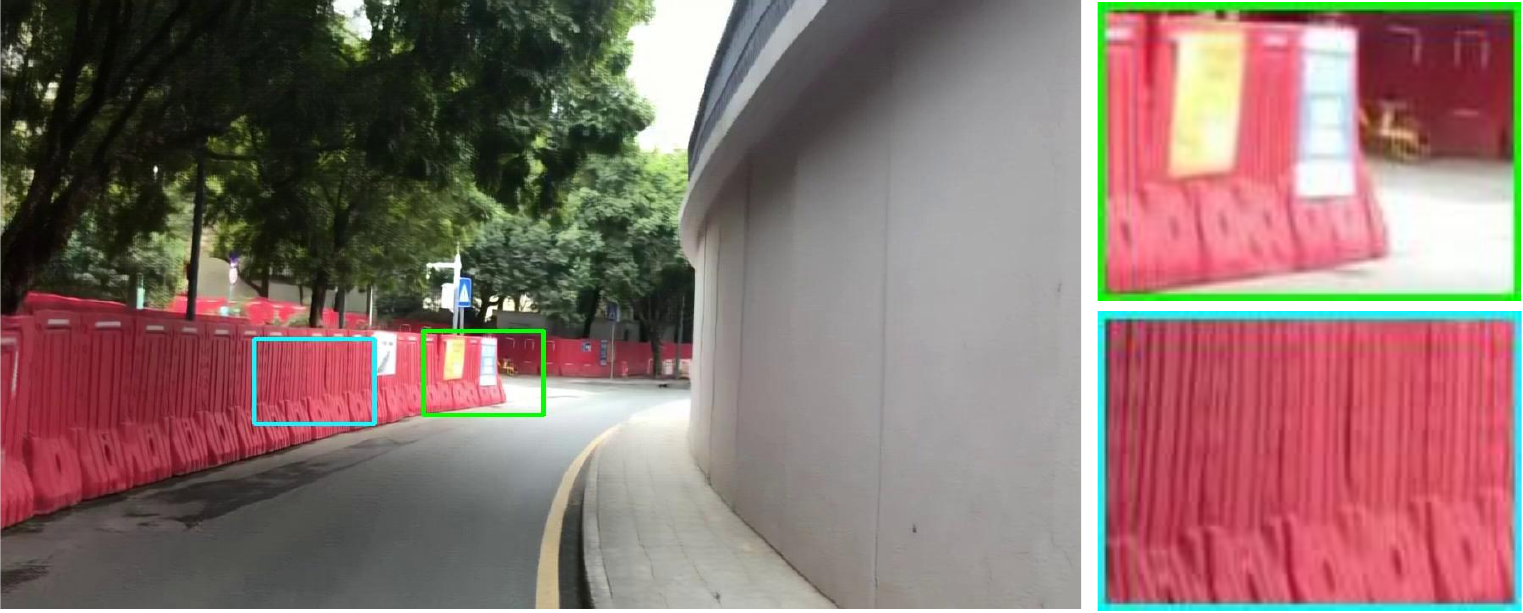}
\\
& STFAN~\cite{zhou2019stfan} & Our MAML & Ours\\

\end{tabular}
\caption{Visual comparisons on our Real-World dataset. Ours is from our fitting-to-test-data pipeline, and our MAML is from our accelerated pipeline. Our method is capable of restoring the text and high-frequency information.}
\label{fig:result_1}
\end{figure*}

\begin{figure*}[t]
\centering
\begin{tabular}{@{}c@{\hspace{1mm}}c@{\hspace{1mm}}c@{\hspace{1mm}}c@{}}

&\includegraphics[width=0.32\linewidth]{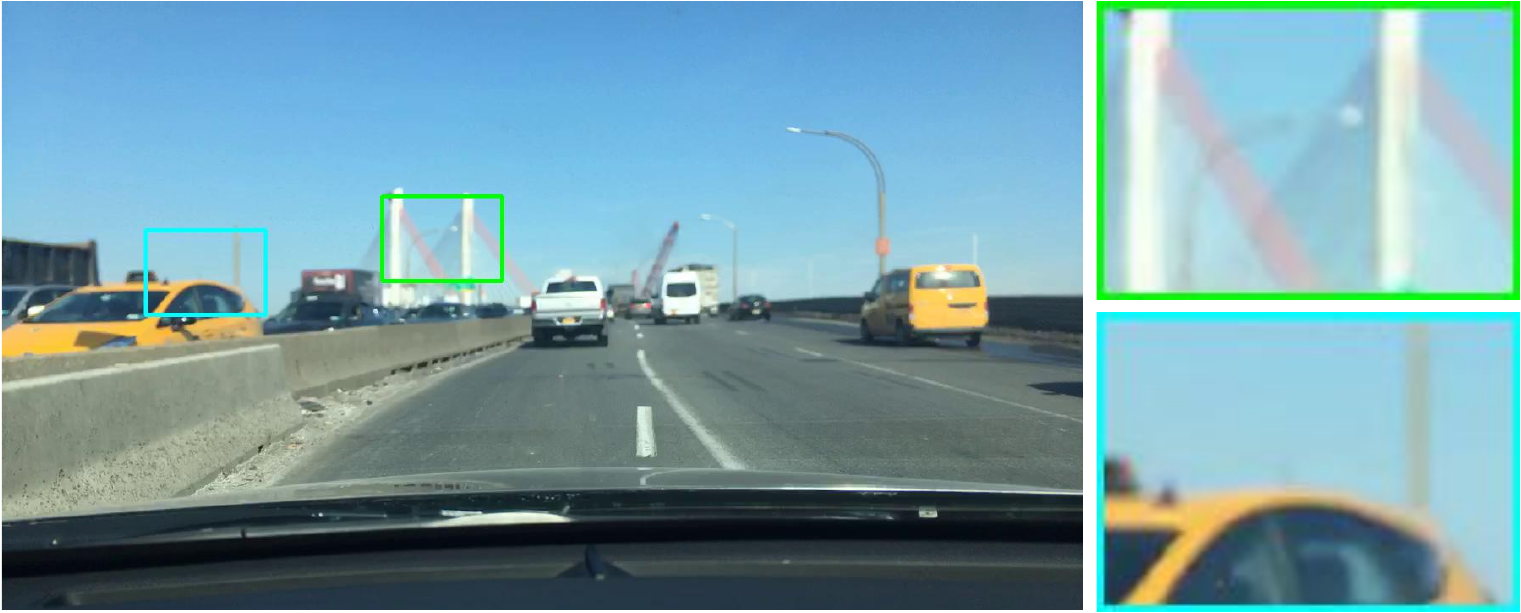}&
\includegraphics[width=0.32\linewidth]{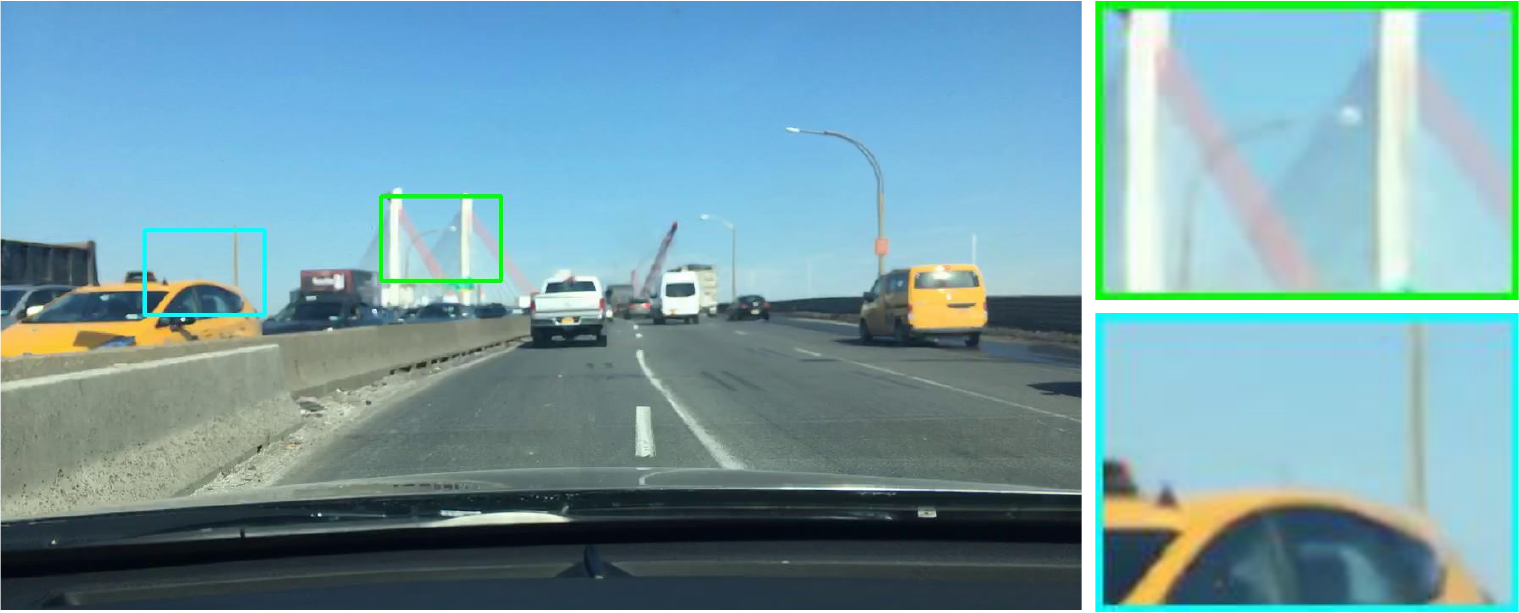}&
\includegraphics[width=0.32\linewidth]{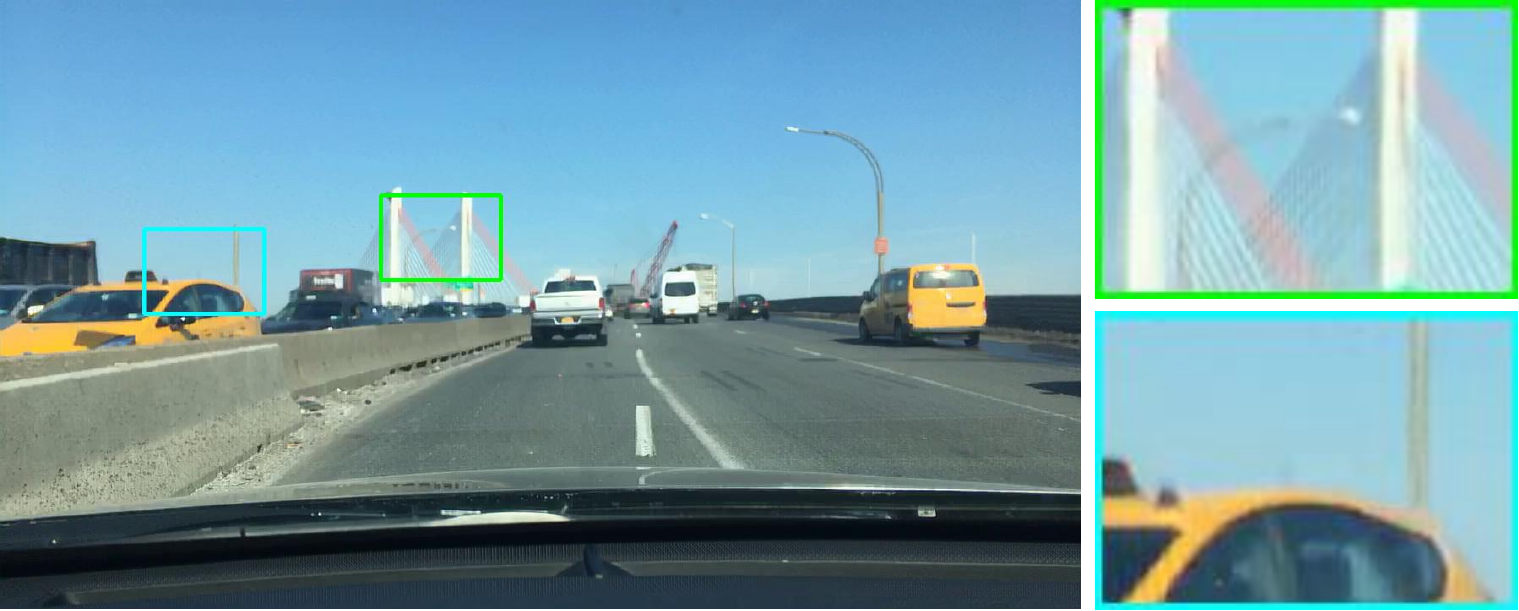}
\\
&\includegraphics[width=0.32\linewidth]{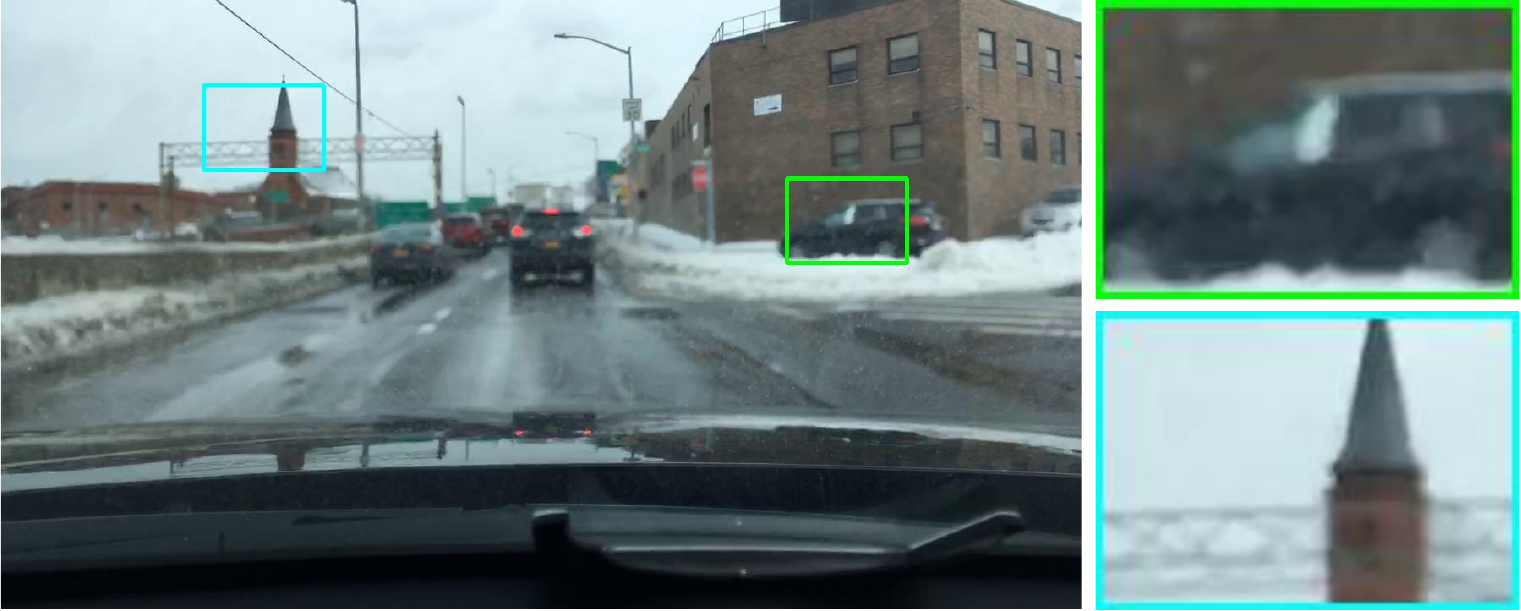}&
\includegraphics[width=0.32\linewidth]{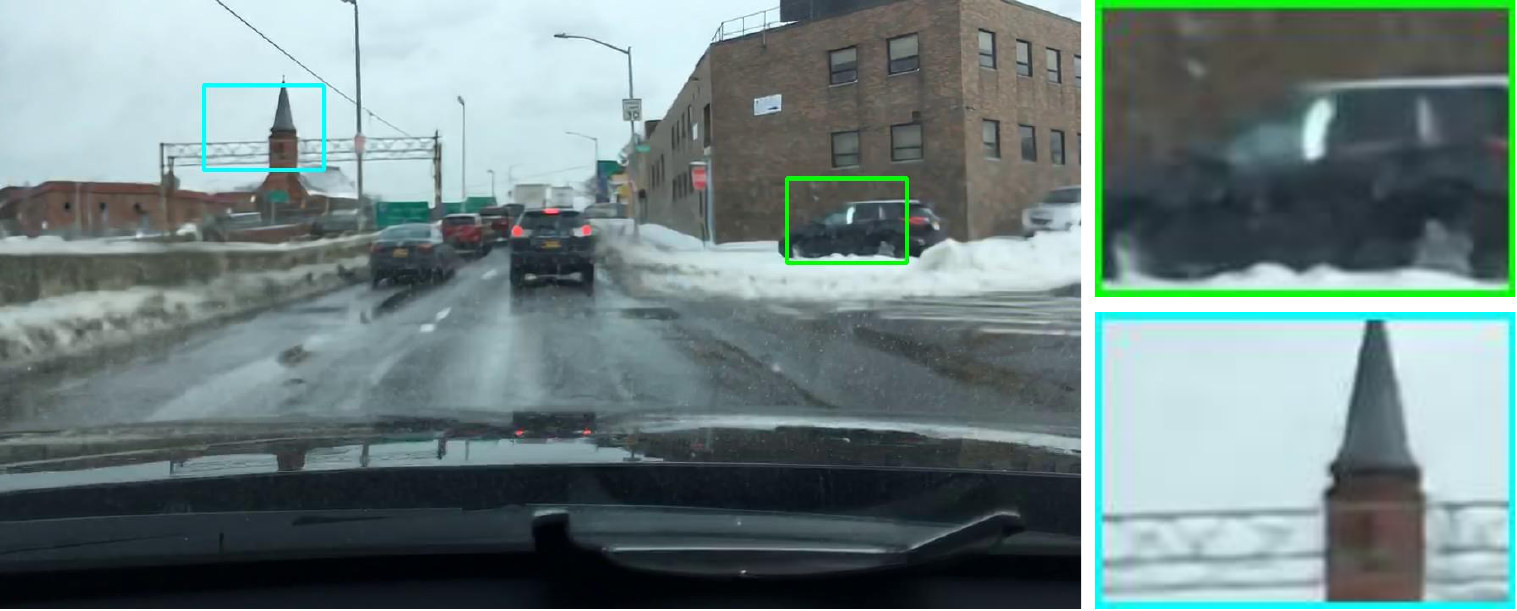}&
\includegraphics[width=0.32\linewidth]{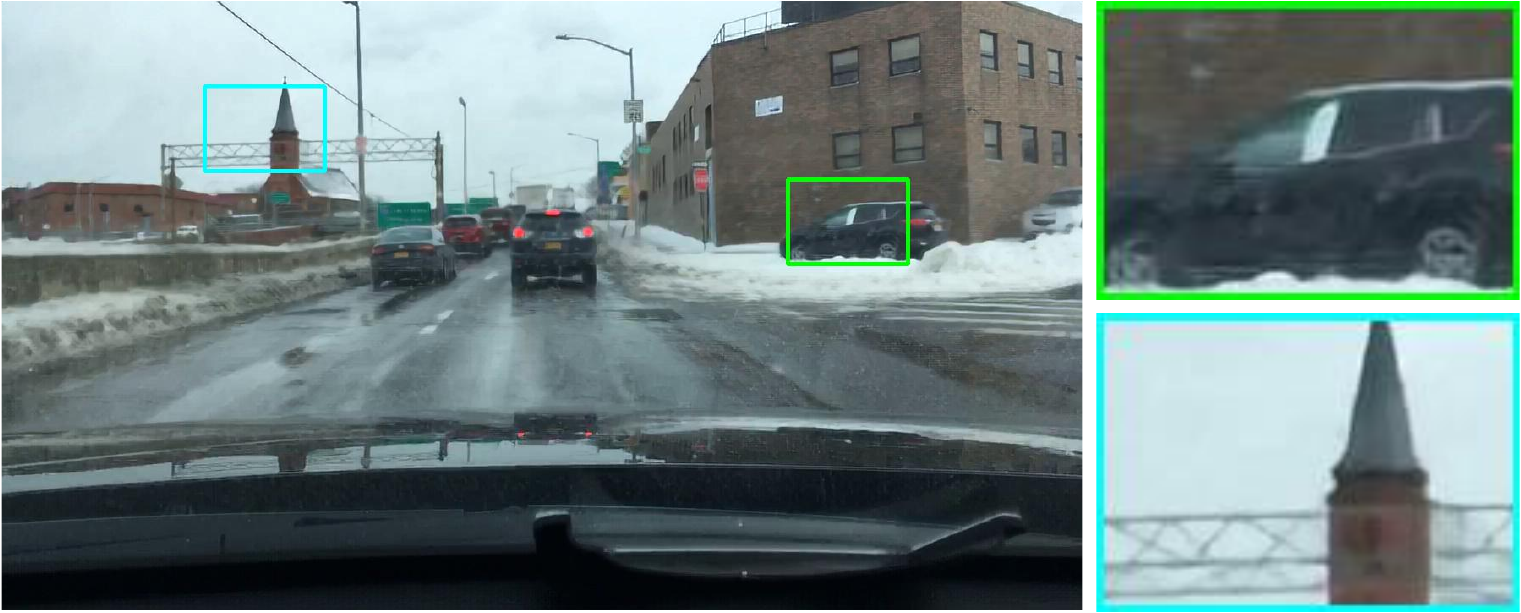}
\\
&Input &   STFAN~\cite{zhou2019stfan} & Ours \\
\end{tabular}
\caption{Visual comparisons on BDD100K dataset. Our pipeline can be well generalized to self-driving scenarios. }
\label{fig:result_bdd}
\end{figure*}

All the results are statistically significant with $p < \num{1e-3}$, and 30 participants are involved in each comparison. Figure~\ref{fig:result_1} and Figure~\ref{fig:result_bdd} display visual comparison examples on different scenarios. 

\begin{figure*}[t]
\centering
\begin{tabular}{@{}c@{\hspace{1mm}}c@{\hspace{1mm}}c@{\hspace{1mm}}c}
\includegraphics[width=0.23\linewidth]{{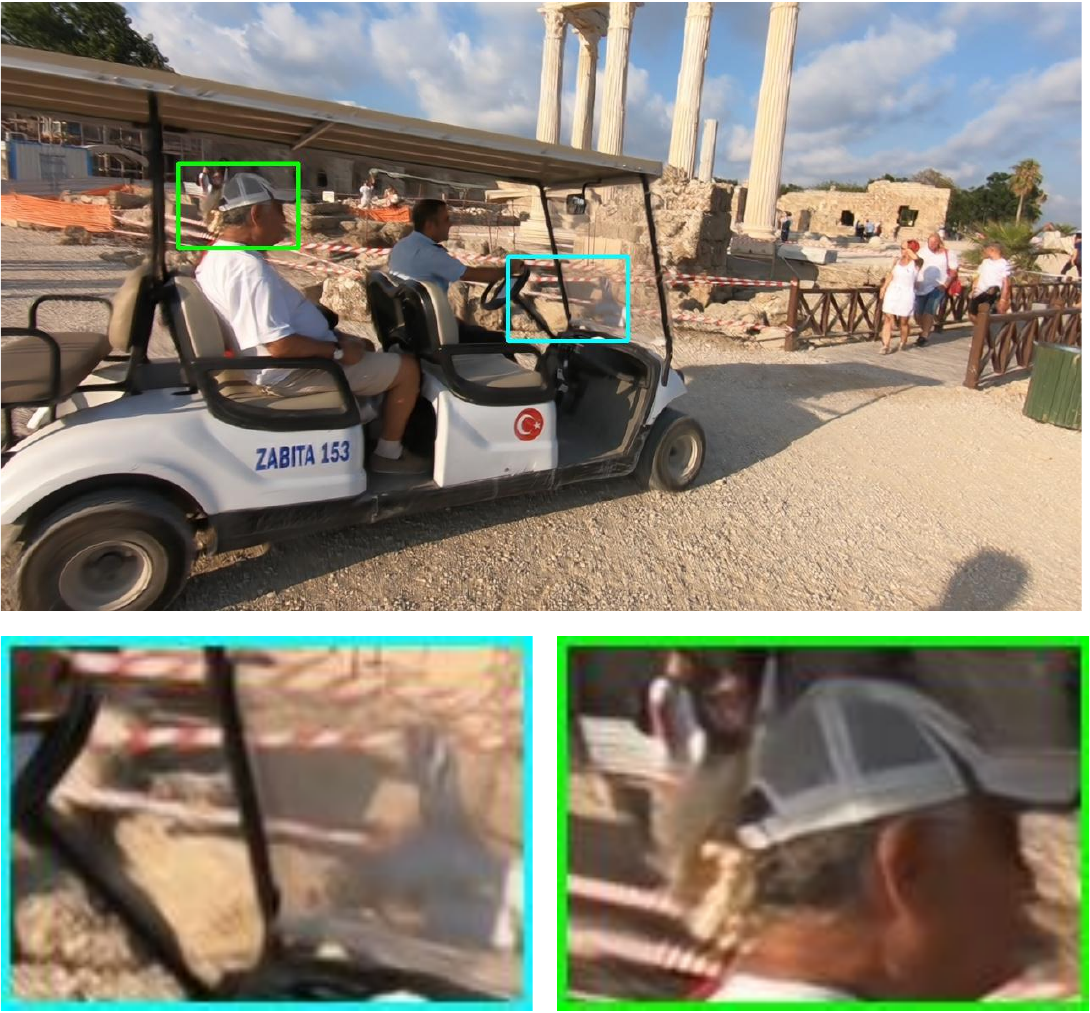}} &
\includegraphics[width=0.23\linewidth]{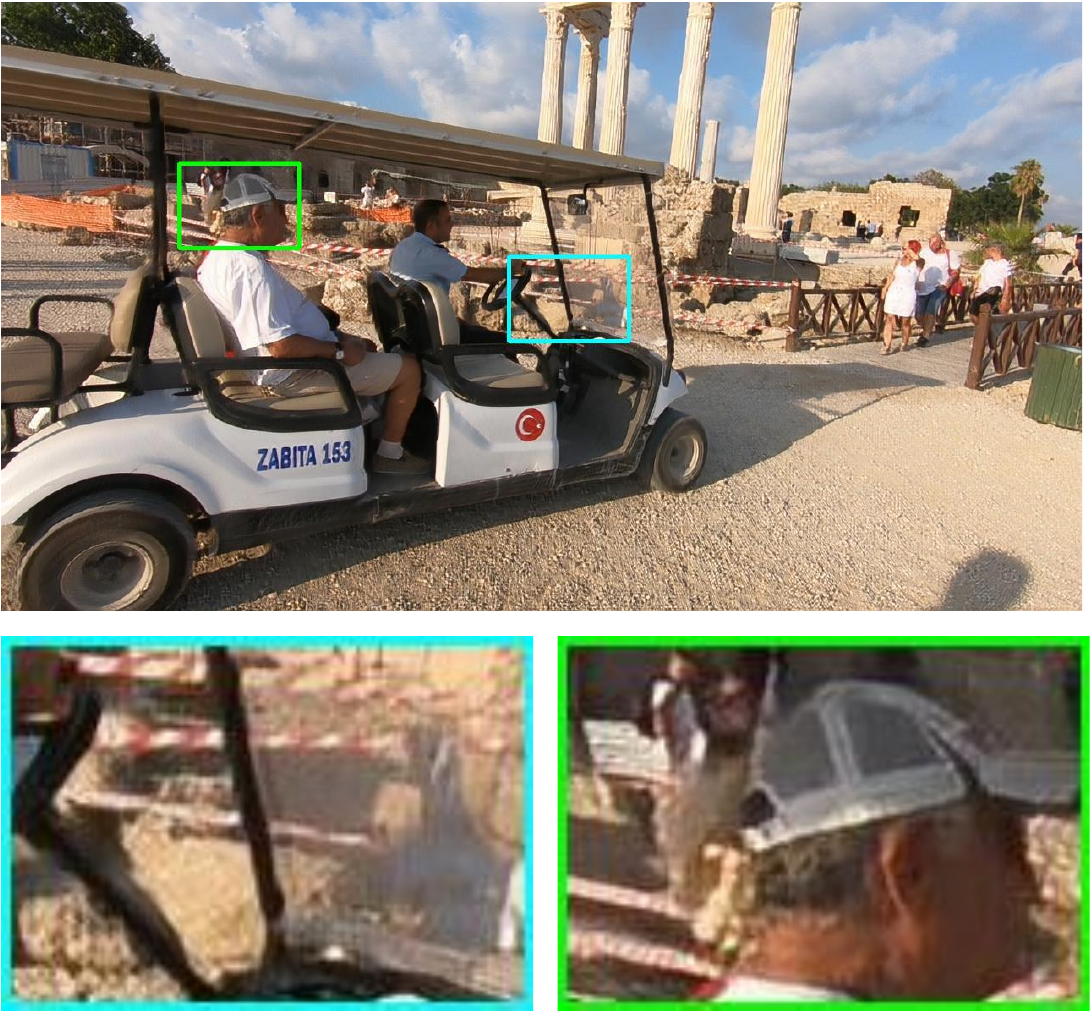} &
\includegraphics[width=0.23\linewidth]{{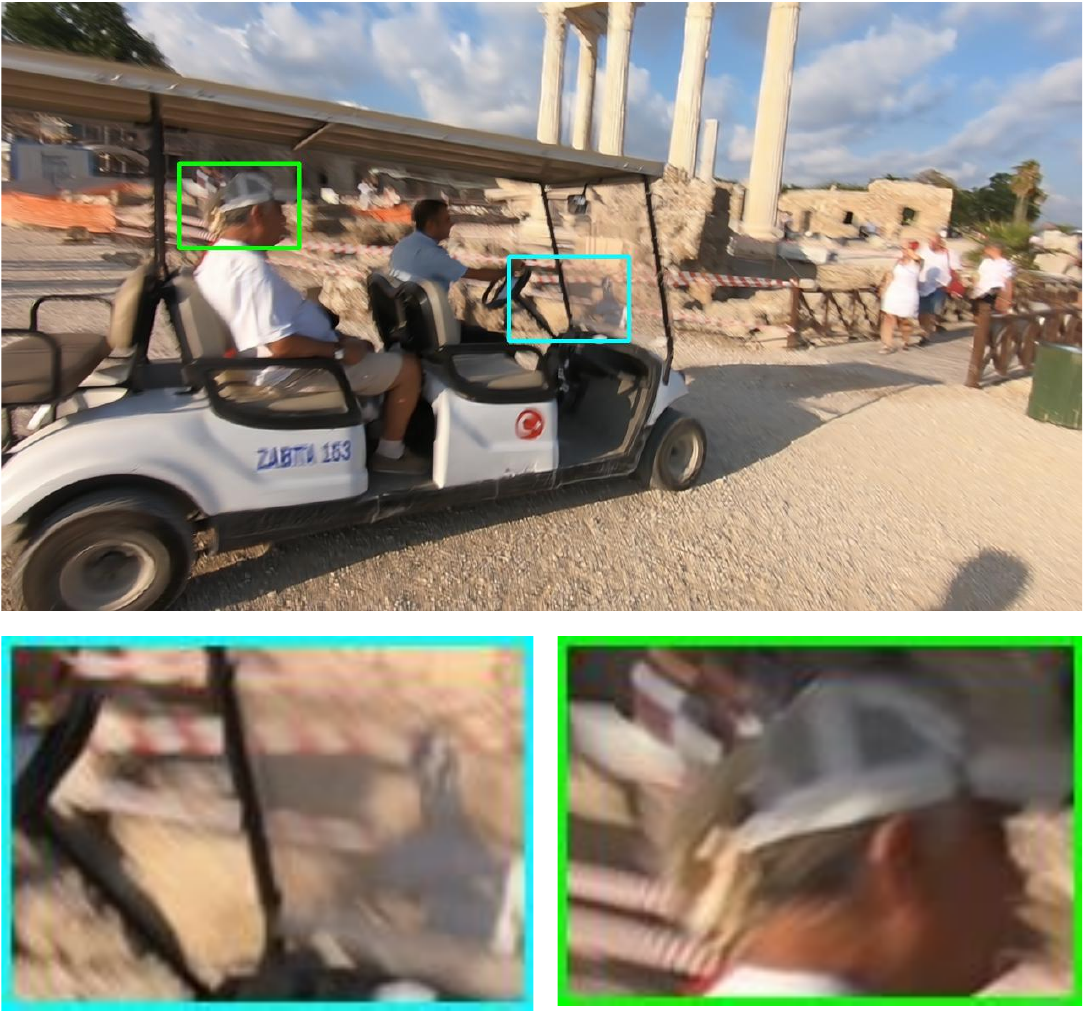}} &
\includegraphics[width=0.23\linewidth]{{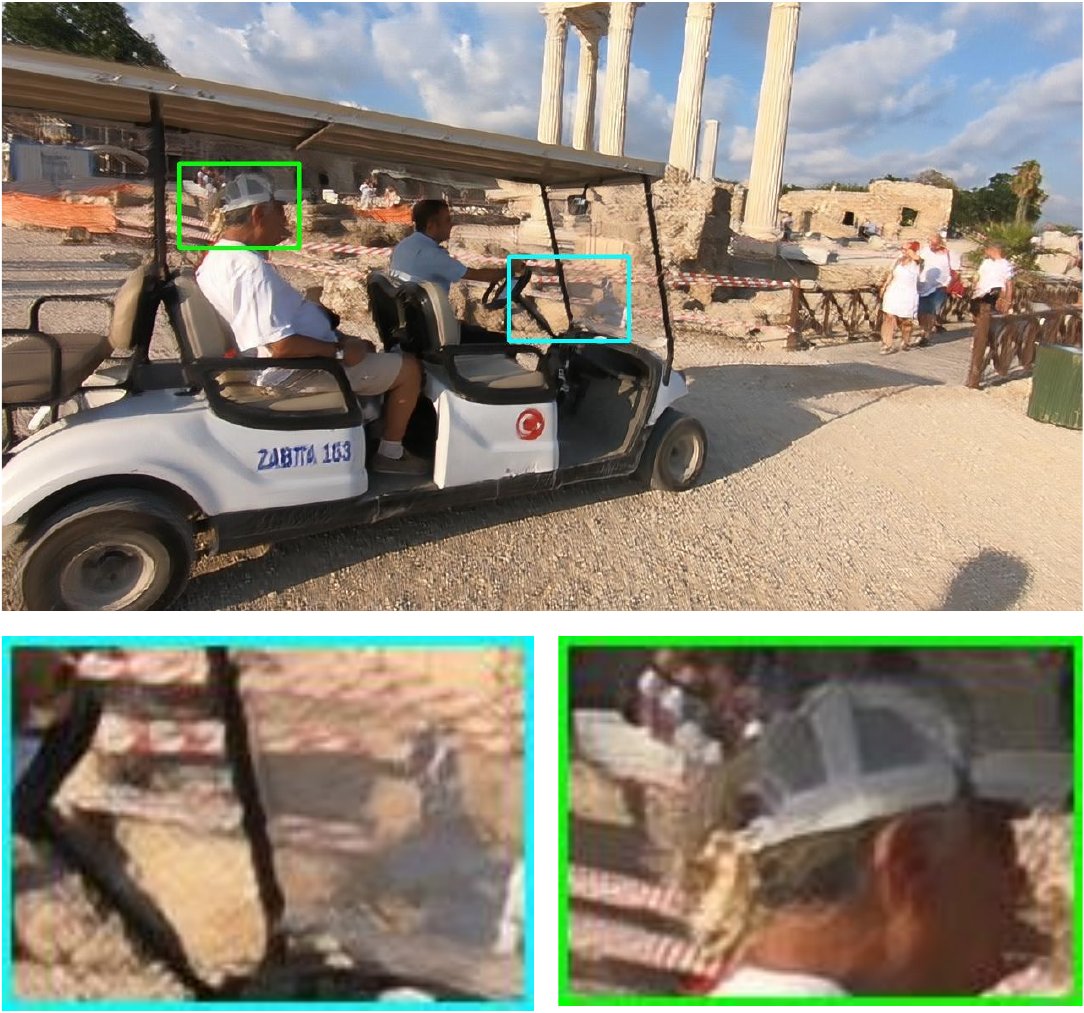}}\\
EDVR~\cite{wang2019edvr} & Our post-processing & STFAN~\cite{zhou2019stfan} & Our post-processing \\
\end{tabular}
\caption{Visual comparison between the results produced by baselines and the post-processing results by our pipeline.}
\label{fig:post}
\end{figure*}

\subsection{Temporal consistency}
We evaluate the temporal consistency by estimating the warping error $E_{warp}$ proposed in ~\cite{lei2020dvp}. We compare the temporal consistency of our method with DMPHN~\cite{Zhang_2019_CVPR}, EDVR~\cite{wang2019edvr}, STFAN~\cite{zhou2019stfan} and DeblurGAN-v2~\cite{kupyn2019deblurgan}. 
As shown in Table~\ref{tbl:Temporal}, our method have the lowest warping error and thus best temporal consistency, though our method is on image level. 
For this metric, We random sample 10 videos from the synthetic datasets~\cite{Nah_2019_CVPR_Workshops_REDS, su2017deep}. 
Please refer to the supplementary video for further demonstration of the temporal consistency.

\subsection{Ablation study}

\begin{table}[t]
\centering
\caption{User study results for ablation study, showing the preference rate of our complete model vs. an ablated model. The images attained with reweighting, reversed gamma correction, and symmetric blur kernels are preferred by users.}
\begin{tabular}{l@{\hspace{5mm}}c}
\toprule
Comparison & Ours \\
\midrule  
Ours without reweighting & 66.0\% \\
Ours without reversed gamma & 83.7\% \\
Ours with simulated kernels & 61.7\% \\
Ours with asymmetric kernels & 84.0\% \\
\bottomrule
\end{tabular}
\label{tbl:ablation}
\vspace{-1em}
\end{table}

In addition to the qualitative comparison for blur kernels shown in Figure~\ref{fig:comparison_kernel}, we also perform an ablation study for different components of our fitting-to-test-data pipeline. As shown in Table~\ref{tbl:Temporal}, we show our proposed symmetric kernels can better improve temporal stability, compared to asymmetric and simulated kernels.

We then show that our reweighting strategy for the loss function, the reverse of gamma correction in blur generation, and symmetric kernels can further improve the performance. We conduct an additional user study on Amazon Mechanical Turk following the above protocol to evaluate the influence brought by these components in our pipeline. As summarized in Table~\ref{tbl:ablation}, users prefer the results synthesized by the pipeline with our full model. For each comparison, we randomly sample 10 images from our real-world video deblurring dataset, and 15 participants are involved in the user study.

\begin{table}[t]
\caption{Comparison between original results by baselines and the post-processing results by our pipeline.}\label{tbl:extension}
\begin{subtable}[t]{\linewidth}  
\centering
\begin{tabular}{l@{\hspace{4em}}c}
\toprule
Comparison & Our post-processing \\
\midrule
EDVR~\cite{wang2019edvr} & 73.3\% \\
STFAN~\cite{zhou2019stfan} & 86.0\% \\
\bottomrule
\end{tabular}
\vspace{0.5em}
\caption{Results of user study.}\label{tbl:usr}
\end{subtable}
\begin{subtable}[t] {\linewidth}
\centering
\begin{tabular}{l@{\hspace{1em}}c@{\hspace{1em}}c}
\toprule
Method & Original & Our post-processing \\
\midrule
EDVR~\cite{wang2019edvr} & {0.2779} & {0.2681} \\
STFAN~\cite{zhou2019stfan} & {0.2716} & {0.2630} \\
\bottomrule
\end{tabular}  
\vspace{0.5em}
\caption{Results of temporal consistency.}\label{tbl:temporal}
\end{subtable}
\end{table}

\subsection{Extension}

In an extreme case, when there are no sharp frames in the test video, our method can also serve as a post-processing step for an existing video deblurring model. We can first apply a pretrained video deblurring model and then use the resulted video as the input to our pipeline for further refinement. We use our pipeline to refine the results by EDVR~\cite{wang2019edvr} and STFAN~\cite{zhou2019stfan} and sample 30 videos from the synthetic datasets for comparison. As shown in Table~\ref{tbl:temporal}, the videos become more temporally consistent after our post-processing. The video quality is also enhanced, as demonstrated in Table~\ref{tbl:usr}. The qualitative comparison is shown in Figure~\ref{fig:post}.

\section{Conclusion}
We have demonstrated the effectiveness of our fitting-to-test-data pipeline for video deblurring. Our approach avoids the domain gap issue between training and test by fitting a deep network on the test video only. 
Our approach is also based on a key observation that there are almost always sharp frames in a blurry video so that we can train a deep model on those sharper video frames. To evaluate our method and prior work on real-world data for video deblurring, we collected a dataset containing videos with motion blur, and the dataset will be released publicly. Furthermore, we have applied meta-learning to accelerate our pipeline significantly, while the quality of deblurred videos only degrades a little bit. To further improve our approach in the future, joint learning of realistic blur kernels in our model is a promising direction.
We hope our fitting-to-test-data pipeline can inspire more researchers to tackle image and video processing with a new paradigm. 

\bibliographystyle{IEEEtran}
\bibliography{root}

\end{document}